\documentclass[11pt, a4paper, logo, copyright, nonumbering]{map}

\usepackage{hyperref}
\usepackage{url}
\usepackage{enumitem}
\usepackage{graphicx} 
\usepackage{booktabs}
\usepackage{multirow}
\usepackage{pifont}
\usepackage{xcolor}
\usepackage{graphicx}
\usepackage{subcaption}
\usepackage{lipsum}
\usepackage{caption}
\usepackage{wrapfig}
\usepackage[most]{tcolorbox}
\usepackage{fontawesome}

\usepackage{listings}
\usepackage{tcolorbox}
\usepackage{graphicx}
\usepackage{caption}
\usepackage{makecell}
\tcbuselibrary{skins,breakable,listings}
\usepackage{algorithm}
\usepackage{algorithmic}
\usepackage{float}
\usepackage{subcaption}
\usepackage{multicol}
\usepackage[table]{xcolor}

\definecolor{cvprblue}{rgb}{0.21,0.49,0.74}
\definecolor{MidnightBlue}{RGB}{25,25,112}

\usepackage{natbib}

\usepackage{CJKutf8}

\usepackage{xargs}  

\usepackage{todonotes}  

\usepackage{multirow}

\usepackage{cleveref}

\usepackage{amsmath}
\usepackage{dsfont}

\usepackage{subcaption}

\usepackage{svg}
\PassOptionsToPackage{most}{tcolorbox}  
\usepackage{tcolorbox}  
\usepackage{fontawesome}

\usepackage[utf8]{inputenc}
\usepackage{booktabs}
\usepackage{graphicx}
\usepackage{array}
\usepackage{tabularx}
\usepackage{xcolor,colortbl}  
\definecolor{boxcolor}{HTML}{d92523} 
\definecolor{bulbcolor}{HTML}{e3b87f} 

\hypersetup{
    colorlinks=true,            
    linkcolor=MidnightBlue,     
    filecolor=magenta,          
    urlcolor=cyan,              
    citecolor=cvprblue,         
    pdftitle={Overleaf Example},
    pdfpagemode=FullScreen,
}
 
\renewcommand{\today}{}
\newcommandx{\info}[2][1=]{\todo[linecolor=red,backgroundcolor=red!25,bordercolor=red,#1]{#2}}

\title{OProver: A Unified Framework for Agentic Formal Theorem Proving}

\author{
\textbf{M-A-P}
}

\begin{abstract}
Recent progress in formal theorem proving has benefited from large-scale proof generation and verifier-aware training, but agentic proving is rarely integrated into prover training, appearing only at inference time. We present OProver, a unified framework for agentic formal theorem proving in Lean~4, in which failed proof attempts are iteratively revised using retrieved compiler-verified proofs and Lean compiler feedback. OProver is trained through continued pretraining followed by iterative post-training: each iteration runs agentic proving, indexes newly verified proofs into OProofs and the retrieval memory, uses repair trajectories as SFT data, and uses unresolved hard cases for RL. OProofs is built from public Lean resources, large-scale proof synthesis, and agentic proving traces, containing 1.77M Lean statements, 6.86M compiler-verified proofs, and serialized trajectories with retrieved context, failed attempts, feedback, and repairs. Across five benchmarks, OProver-32B attains the best Pass@32 on MiniF2F (93.3\%), ProverBench (58.2\%), and PutnamBench (11.3\%), and ranks second on MathOlympiad (22.8\%) and ProofNet (33.2\%)---more top placements than any prior open-weight whole-proof prover. 
\end{abstract}

\makeatletter
\renewcommand{\abscontent}{
  \begin{center}
  \small
  \raisebox{-1.2pt}{\includegraphics[height=9pt]{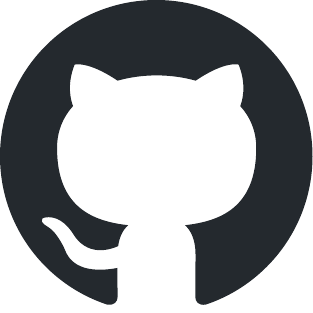}}\,
  \href{https://github.com/multimodal-art-projection/OProver}{\texttt{multimodal-art-projection/OProver}}
  \quad
  \raisebox{-1.2pt}{\includegraphics[height=9pt]{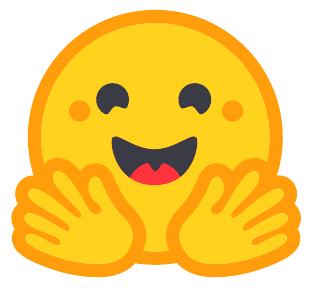}}\,
  \href{https://huggingface.co/collections/m-a-p/oprover}{\texttt{huggingface.co/m-a-p/oprover}}
  \end{center}
  \vspace{1em}
  \noindent
  \centerline{\fontsize{15pt}{14pt}\selectfont\textbf{Abstract}}\vspace{5ex}
  \parbox{\dimexpr\linewidth}{\absfont \theabstract}
  \@ifundefined{@keywords}{}{
      \vskip1em \noindent \keywordsfont  Keywords: \@keywords}
}
\makeatother

\begin{document}
\maketitle
\begin{CJK*}{UTF8}{gbsn}

\begin{figure}[H]
\centering
\includegraphics[width=\textwidth]{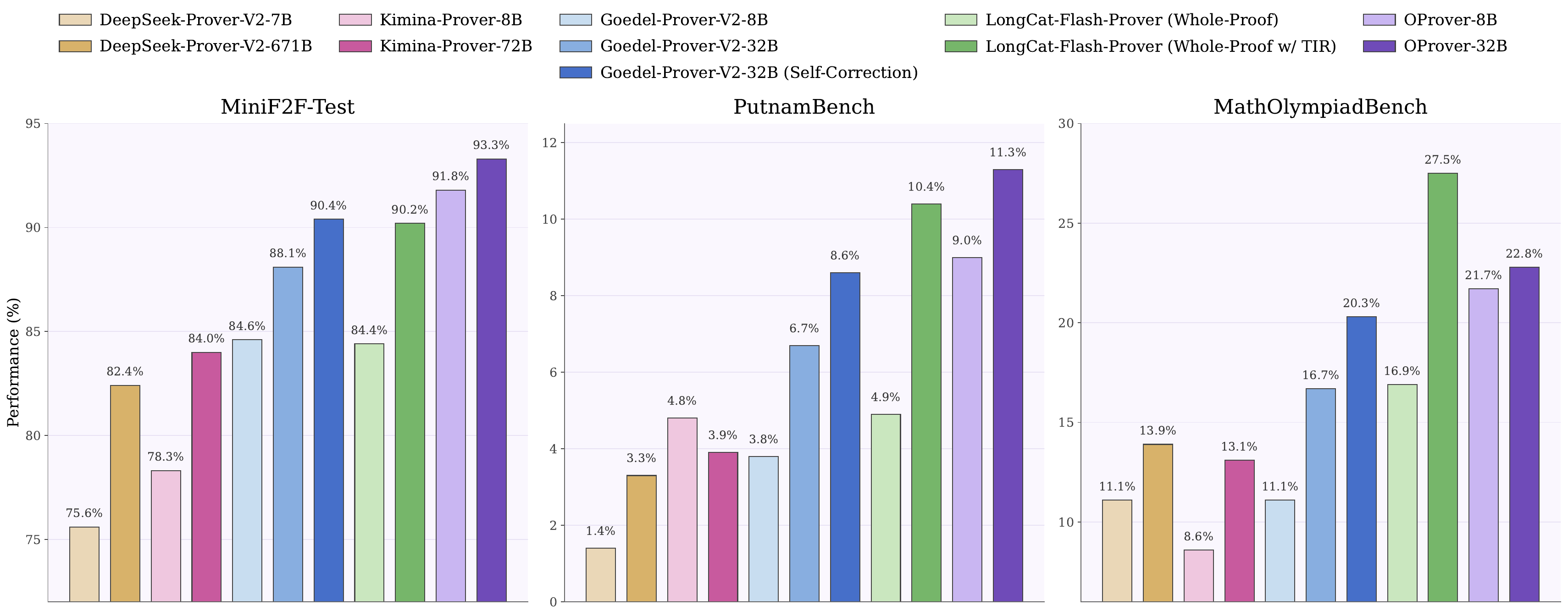}
\caption{Pass@32 performance on MiniF2F-Test, PutnamBench, and MathOlympiadBench. OProver consistently outperforms prior open-source whole-proof provers across all three benchmarks; the only exception is LongCat-Flash-Prover with TIR on MathOlympiadBench.}
\label{fig:main_benchmark_comparison}
\end{figure}

\section{Introduction}

Formal theorem proving in systems such as Lean~4~\citep{moura2021lean4theoremprover} provides a rigorous setting for machine reasoning, where every step of a proof is mechanically verified by a small, trusted kernel. This makes them a natural foundation for reliable mathematical reasoning and verified software, but also imposes a high bar: a proof is accepted only if it is fully formal and type-correct. Recent prover systems built on Lean have substantially improved performance on challenging benchmarks such as MiniF2F~\citep{zheng2021minif2f} and PutnamBench~\citep{tsoukalas2024putnambench}, yet absolute success rates on harder benchmarks remain low, and most systems still rely primarily on single-pass or best-of-$N$ whole-proof generation. Retrieval and compiler feedback, when used at all, are typically applied as test-time heuristics rather than as a proving policy that the prover is trained to use, leaving an important source of supervision largely unexploited.

A small but growing line of work has begun to incorporate retrieval, compiler feedback, and iterative repair into formal theorem proving~\citep{yang2023leandojo, jiang2022draft, wu2025internlm2}. However, these capabilities are typically introduced as inference-time augmentations on top of a fixed prover, rather than as a learned policy that the model is trained to use. As a result, a prover trained mainly on finalized proofs sees compiler feedback and retrieved evidence only at deployment, in distributions it was never optimized for. Closing this train--inference mismatch requires training the prover to perform retrieval-grounded, feedback-conditioned refinement as part of its policy, not as a separate inference-time procedure.

Training such a policy in turn requires data that existing formal corpora do not provide. Public formal theorem proving corpora and proof-synthesis datasets~\citep{wu2022autoformalization,ying2024lean,peng2025criticlean} focus on the end products of proving: large collections of formal statements and final compiler-verified proofs. In general, they omit failed attempts, retrieved context, and compiler diagnostics that drive proof repair in practice. Autoformalization and proof synthesis have greatly expanded the number of available statements and verified proofs, but they do not by themselves record the multi-round interaction histories needed to learn agentic self-correction. What is missing is therefore not more verified proofs, but corpora that explicitly preserve how proofs are constructed, fail, and get repaired.

In this paper, we present \textbf{OProver}, a unified framework for agentic formal theorem proving in Lean~4. Rather than treating retrieval, compiler feedback, and iterative repair as separate inference-time modules built on top of a fixed prover, OProver unifies them with training and data construction into a single proving framework. At inference time, OProver treats proving as a multi-round refinement loop, in which each failed proof attempt is revised using retrieved compiler-verified proofs and Lean~4 compiler feedback. At training time, the same retrieval and feedback signals shape the prover's policy: training proceeds in two phases, continued pretraining on Lean code and mathematics, followed by iterative post-training in which agentic proving, supervised fine-tuning, and reinforcement learning alternate. Newly verified proofs and repair traces produced by the current prover are recirculated into OProofs and the retrieval memory, so that the data, the training procedure, and the proving policy co-evolve within a single framework.

We pair OProver with \textbf{OProofs}, a large-scale corpus for agentic formal theorem proving that supports both pretraining and the co-evolution loop above. Unlike prior Lean datasets, which mostly preserve only final compiler-verified proofs, OProofs additionally records the trajectories of proof construction, failure, feedback, and repair. It contains 1.77M Lean statements and 6.86M compiler-verified proofs, paired with serialized proving trajectories that capture retrieved context, failed attempts, compiler feedback, and subsequent repairs. We build OProofs from three complementary sources: public Lean resources, large-scale proof synthesis with compiler verification, and traces from OProver's own agentic proving. The first two sources provide an initial corpus that supports both pretraining and post-training, while the third grows continuously as OProver improves: newly verified proofs are indexed into the retrieval memory and repair trajectories are added to subsequent training rounds.

In summary, our contributions are:
\begin{enumerate}[leftmargin=*]
\item \textbf{A unified framework for agentic formal theorem proving.} We propose OProver, which treats proving as a retrieval-grounded, feedback-conditioned refinement loop and trains the prover end-to-end to use these signals as part of its policy, rather than as test-time
heuristics.

\item \textbf{A large-scale corpus for agentic formal theorem proving.} We construct OProofs, containing 1.77M Lean statements, 6.86M compiler-verified proofs, and serialized proving trajectories that record retrieved context, failed attempts, compiler feedback, and subsequent repairs---supervision that prior Lean corpora do not provide.

\item \textbf{A co-evolution pipeline between prover and corpus.} We develop an iterative post-training pipeline in which the current prover produces new verified proofs and repair trajectories: verified proofs are indexed into the retrieval memory, repair trajectories become SFT data, and the hardest unresolved cases provide RL signal for the next round.

\item \textbf{State-of-the-art performance among open-weight whole-proof provers.} OProver-32B attains three best and two second-best results across five formal theorem-proving benchmarks, reaching Pass@32 of 93.3 on MiniF2F, 58.2 on ProverBench, and 11.3 on PutnamBench.
\end{enumerate}
\section{Methodology}
\label{sec:oprover}

\begin{figure}[t]
  \centering
  \includegraphics[width=\linewidth]{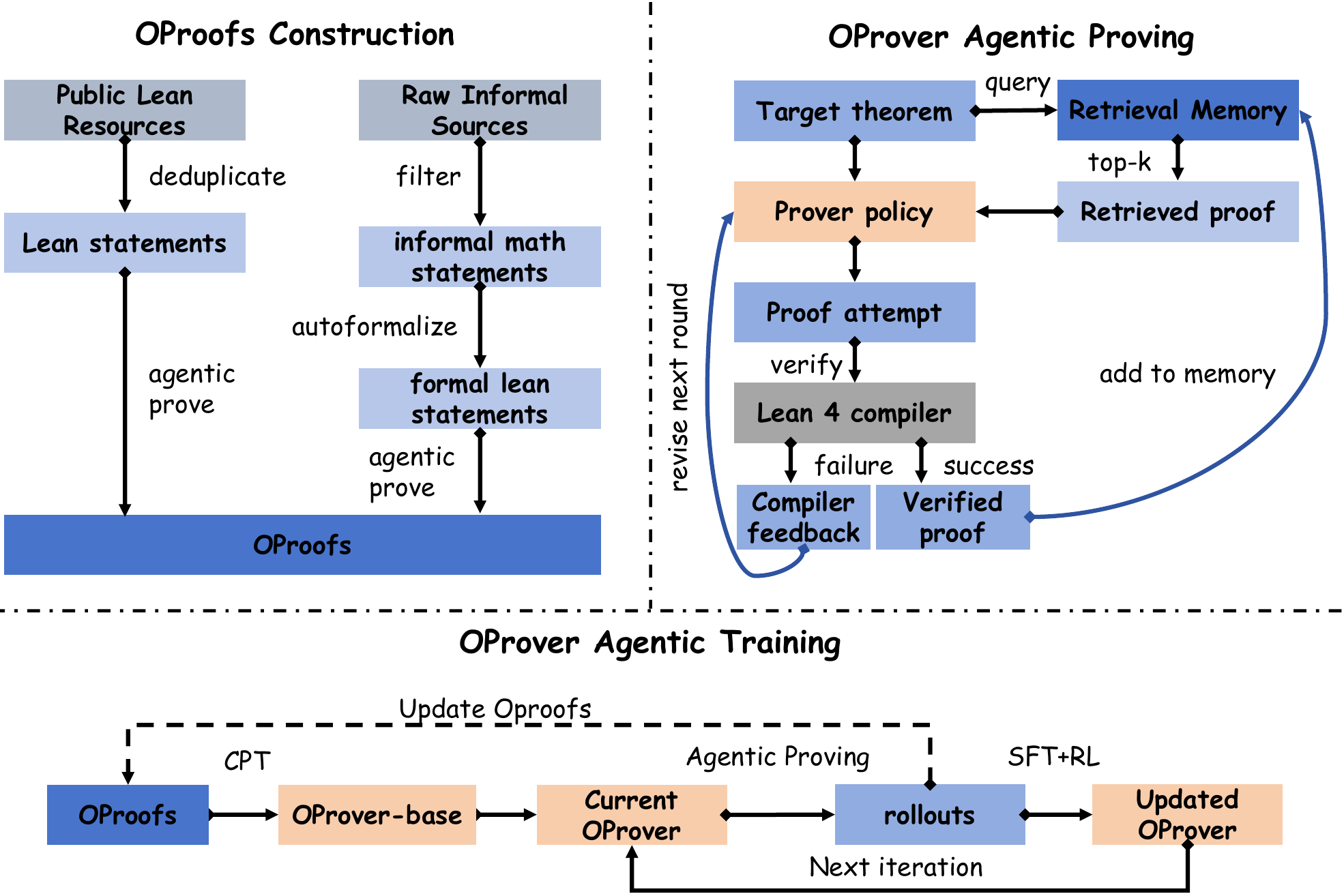}
  \caption{
Overview of OProver. The framework has three components: \textbf{OProofs Construction} (top-left), which builds a Lean-specific corpus from public Lean resources and autoformalized statements; \textbf{OProver Agentic Proving} (top-right), which performs multi-round refinement under retrieval and Lean~4 compiler feedback; and \textbf{OProver Agentic Training} (bottom), where a one-time CPT yields OProver-Base, followed by an iterative post-training loop in which agentic proving, SFT, and RL produce $\text{OProver}_{t+1}$ from $\text{OProver}_t$, while verified proofs are folded back into OProofs.
  }
  \label{fig:oprover_overview}
\end{figure}

Figure~\ref{fig:oprover_overview} illustrates the three components of our framework. The \textbf{OProofs Construction} pipeline (top-left) builds a Lean-specific corpus from public Lean resources and raw informal sources, through deduplication, filtering, autoformalization, agentic proving, and Lean~4 verification (\S\ref{sec:oproofs}). The \textbf{OProver Agentic Proving} pipeline (top-right) performs theorem proving as a multi-round interaction: given a target theorem, the prover policy queries a retrieval memory for top-$k$ compiler-verified proofs, produces a proof attempt, and is verified by the Lean~4 compiler; on failure, compiler feedback is returned to the policy and the proof is revised in the next round, while successful proofs are added back into the retrieval memory (\S\ref{sec:oprover_framework}). The \textbf{OProver Agentic Training} pipeline (bottom) trains OProver in two stages: a one-time continued pretraining on OProofs yields a domain-adapted base model OProver-Base, followed by an iterative post-training loop in which the current prover performs agentic rollouts and is updated by SFT on round-level repair examples and RL on harder unresolved cases; verified proofs and repair trajectories from each iteration are folded back into OProofs and the retrieval memory, so that the corpus and the prover co-evolve across post-training iterations (\S\ref{sec:training}).


\subsection{The OProver Framework}
\label{sec:oprover_framework}

OProver consists of three interacting components: a proving policy $\pi$, a retrieval memory $\mathcal{M}$ of compiler-verified proofs, and the Lean~4 compiler $\mathcal{V}$ as the verification environment. The policy is trained to use the other two components through the agentic proving formulation below.

\paragraph{Agentic proving formulation.}
We formulate theorem proving as a bounded multi-round refinement process. At round $t$, the policy conditions on a state
\[
X_t = (s,\; \mathcal{R}_t,\; p_{t-1},\; f_{t-1}),
\]
where $s$ is the target theorem statement, $\mathcal{R}_t$ is the retrieved proof context, $p_{t-1}$ is the previous proof attempt, and $f_{t-1}$ is the corresponding compiler feedback. The policy then produces a revised proof attempt
\[
p_t \sim \pi(\cdot \mid X_t).
\]
At the initial round, $p_0$ and $f_0$ are empty. The interaction terminates when $\mathcal{V}$ verifies a proof attempt or when a predefined round budget $T$ is exhausted.

\paragraph{Compact interaction state.}
A key modeling choice in OProver is that the policy conditions only on the \emph{most recent} proof attempt and its compiler feedback, rather than on the full interaction history $(p_0, f_0, \ldots, p_{t-1}, f_{t-1})$. This keeps the interaction state compact and preserves the local correction signal most relevant for proof repair. Crucially, the same state formulation is used throughout rollout collection, supervised fine-tuning, and reinforcement learning, so that the training-time interface exactly matches the proving-time interaction (\S\ref{sec:training}).

\paragraph{Retrieved compiler-verified proofs.}
At each round, OProver retrieves top-$k$ compiler-verified proofs from the retrieval memory $\mathcal{M}$ by semantic similarity to the target statement~$s$, using a sentence-embedding model trained on Lean theorem--proof pairs (details in \S\ref{sec:oproofs}). $\mathcal{M}$ is constructed from OProofs and is continuously expanded with newly verified proofs during iterative post-training, so the retrieval context becomes richer as training progresses. The retrieved proofs provide reusable lemma usage patterns, tactic structures, and proof strategies from related formal contexts, and are inserted into the policy input alongside~$s$, $p_{t-1}$, and $f_{t-1}$.

\paragraph{Compiler feedback as a correction signal.}
When a proof attempt $p_{t-1}$ fails Lean verification, $\mathcal{V}$ returns textual diagnostics describing the failure---syntax errors, type mismatches, unknown identifiers, tactic failures, or unsolved goals. We pass these diagnostics directly to the policy as $f_{t-1}$, without projecting them into a hand-designed error taxonomy. Preserving the raw textual form lets the policy condition on fine-grained information that would be lost under categorical encoding, and is essential for the targeted revisions that distinguish multi-round refinement from repeated single-shot regeneration.

\subsection{OProofs: A Unified Training Corpus}
\label{sec:oproofs}

OProofs is a Lean-specific corpus that serves two roles in the OProver framework: it provides training data across CPT, SFT, and RL, and it supplies the compiler-verified proofs that populate the retrieval memory $\mathcal{M}$ used at proving time (\S\ref{sec:oprover_framework}). In total, OProofs contains $\sim$1.5M Lean statements and $\sim$5.0M compiler-verified proofs, together with serialized proving trajectories that record retrieved context, failed attempts, compiler feedback, and subsequent repairs.

\paragraph{Construction pipeline.}
Figure~\ref{fig:oprover_overview} (top-left) shows the two-branch construction pipeline, both ending in Lean~4 verification.

\textit{Branch~1: public Lean resources.} We collect open-source Lean statements from \{NuminaMath-LEAN, Lean-Workbook, Leanabell-FormalStmt, Goedel-Pset, \ldots\} and deduplicate at the statement level by exact and near-duplicate matching of theorem signatures. We then run agentic proving with open-source provers on each unique statement and retain only Lean-verified proofs.

\textit{Branch~2: raw informal sources.} We mine informal mathematical statements from Common Crawl and GitHub code repositories using a FastText classifier trained on math-tagged web pages. The filtered statements are autoformalized into Lean~4 statements by Criticlean \cite{peng2025criticlean}, and the same agentic proving process as in Branch~1 is used to obtain Lean-verified proofs.

\paragraph{Round-level repair data.}
Although OProver interacts with Lean over multiple rounds, training does not consume the full interaction history as a single long sequence. Instead, each rollout is decomposed into round-level repair examples of the form $(s, \mathcal{R}_t, p_{t-1}, f_{t-1}) \rightarrow p_t$, matching the state formulation in \S\ref{sec:oprover_framework} exactly. This provides process-level supervision across multi-round rollouts while keeping the training interface identical to the proving-time interaction. The round-level repair set is used by SFT in \S\ref{sec:training}.

\paragraph{An evolving corpus.}
OProofs is not a static dataset. As OProver improves, newly verified proofs and repair trajectories are continuously added back, expanding both the training pool and the retrieval memory $\mathcal{M}$ (details in \S\ref{sec:training}).

\subsection{Training OProver with OProofs}
\label{sec:training}

Training proceeds in two phases: a one-time continued pretraining on OProofs produces a domain-adapted base model OProver-Base, after which iterative post-training alternates agentic proving rollouts with SFT and RL, expanding OProofs as the prover improves (Algorithm~\ref{alg:oprover-oproofs-coevolution}).

\paragraph{Continued pretraining.}
We perform a one-time continued pretraining on a 65B-token mixture designed to strengthen formal reasoning, code, and long-context mathematical reasoning. The mixture contains approximately 30\% Lean-related formal data drawn from OProofs (\S\ref{sec:oproofs}), 20\% code data from OpenCoder~\citep{huang2025opencoder}, 40\% mathematical data from Nemotron-Math-4-Plus~\citep{mahabadi2025nemotron}, and 10\% long-chain-of-thought data from ProLong-64K~\citep{gao2025train}. After CPT, we obtain \emph{OProver-Base}, which serves as the starting point of iterative post-training.

\begin{algorithm}[t]
\caption{Iterative co-evolution of OProver and OProofs}
\label{alg:oprover-oproofs-coevolution}
\begin{algorithmic}[1]
\REQUIRE
CPT-initialized policy $\pi_0$ (OProver-Base);
initial corpus $\mathcal{D}_0$;
theorem pool $\mathcal{Q}$;
Lean verifier $\mathcal{V}$;
number of co-evolution iterations $K$
\STATE $\mathcal{M}_0 \gets \mathrm{Index}(\mathcal{D}_0)$
\FOR{$k = 0, 1, \ldots, K-1$}
  \STATE Run multi-round agentic proving rollouts of $\pi_k$ on $\mathcal{Q}$, retrieving from
$\mathcal{M}_k$ and verifying with $\mathcal{V}$
  \STATE Collect newly verified theorem--proof pairs $\mathcal{P}_k^{+}$
  \STATE Extract round-level repair examples $\mathcal{B}_k$ (each of form $(s, \mathcal{R}_t, p_{t-1}, f_{t-1})
  \rightarrow p_t$, where $t$ is the within-rollout round index)
  \STATE Select hard cases $\mathcal{H}_k$ as groups with non-trivial success rate
  \STATE $\mathcal{D}_{k+1} \gets \mathcal{D}_k \cup \mathcal{P}_k^{+}$;\;\; $\mathcal{M}_{k+1} \gets
\mathrm{Index}(\mathcal{D}_{k+1})$
  \STATE $\pi_{k+1} \gets \mathrm{SFT}(\pi_k;\, \mathcal{B}_k)$ followed by $\mathrm{GSPO}(\,\cdot\,;\,
\mathcal{H}_k,\, \mathcal{M}_{k+1})$
\ENDFOR
\STATE \textbf{return} $\pi_K$, $\mathcal{D}_K$
\end{algorithmic}
\end{algorithm}

\paragraph{Supervised fine-tuning through agentic proving.}
Starting from OProver-Base, we perform agentic proving on theorems that the current model does not yet solve reliably. Each rollout consumes retrieved proof context, prior proof attempts, and compiler feedback, exactly matching the proving-time interaction defined in \S\ref{sec:oprover_framework}. From these rollouts, we extract round-level repair examples $(s, \mathcal{R}_t, p_{t-1}, f_{t-1}) \rightarrow p_t$ and use them as SFT data, computing cross-entropy loss only on the target proof attempt $p_t$. Newly verified proofs are simultaneously added to OProofs and indexed into the retrieval memory $\mathcal{M}$ for subsequent rounds.

\paragraph{Reinforcement learning.}
After SFT, we further improve the policy with Group Sequence Policy Optimization (GSPO)~\citep{zheng2025group}, with group-relative advantage normalization. For each theorem, we sample $n$ multi-round agentic proving rollouts (each containing up to $R$ refinement rounds), and assign a \emph{per-round} reward
\[
r_t =
\begin{cases}
0.8 + 0.2 \cdot \mathbb{1}[\text{format correct}] & \text{if } \mathcal{V}(p_t) = \text{verified},\\
0.0 & \text{otherwise}.
\end{cases}
\]
That is, format quality is rewarded only when the round is verified by Lean. Although rewards are assigned at the round level, all $n \times R$ rounds for the same theorem are pooled into a single group, and advantages are computed by group-relative normalization across this pooled set. This pooled normalization allows the policy to learn from contrasts both across independent attempts and across successive refinement rounds within a single rollout. The state $X_t$ and retrieval memory used during RL are identical to those at proving time (\S\ref{sec:oprover_framework}), so the policy is optimized end-to-end on the same interface it is deployed under. Specific values of $n$, $R$, and other hyperparameters are reported in \S\ref{sec:exp_setup}.

\paragraph{Co-evolution between OProver and OProofs.}
Each post-training iteration follows the loop in Algorithm~\ref{alg:oprover-oproofs-coevolution}. Let $\mathcal{D}_k$, $\mathcal{M}_k$, $\mathcal{B}_k$, and $\mathcal{H}_k$ denote OProofs, the retrieval memory, the SFT repair set, and the RL hard-case set at iteration $k$. The current policy $\pi_k$ runs agentic proving on a theorem pool $\mathcal{Q}$, contributing newly verified proofs $\mathcal{P}_k^{+}$ to $\mathcal{D}_{k+1}$, round-level repair examples to $\mathcal{B}_k$, and groups with non-trivial success rate to $\mathcal{H}_k$. Indexing $\mathcal{D}_{k+1}$ yields the next-iteration retrieval memory $\mathcal{M}_{k+1}$. As $\mathcal{D}_k$ grows across iterations, both retrieval context and supervision signal become richer, jointly driving $\pi_k$ toward stronger proving capability.

\section{Data Statistics}
\label{sec:data_statistics}

\begin{figure*}[t]                                                                                         
\centering
\begin{subfigure}[b]{0.32\linewidth}
\centering                                                                                              
\footnotesize
\setlength{\tabcolsep}{3pt}                                                                             
\begin{tabular}{lc}
  \toprule
  \textbf{Metric} & \textbf{Value} \\
  \midrule
  Unique Lean statements        & 1.77M \\
  Compiler-verified proofs      & 6.86M \\
  \quad with retrieval context  & 4.33M \\                                                              
  \quad with compiler feedback  & 869K  \\
  Agentic proving trajectories  & 1.07M \\                                                              
  \quad with multi-round repair & 164K  \\
  Round-level repair examples   & 280K  \\
  Lean corpus tokens & 51.8B \\
  \bottomrule                                                                                           
\end{tabular}
\vspace{0.3em}                                                                                          
\caption{Corpus overview.}
\label{fig:corpus_overview_table}
\end{subfigure}\hfill                                                                                     
\begin{subfigure}[b]{0.33\linewidth}                                                                      
\centering                                                                                              
\includegraphics[width=0.9\linewidth]{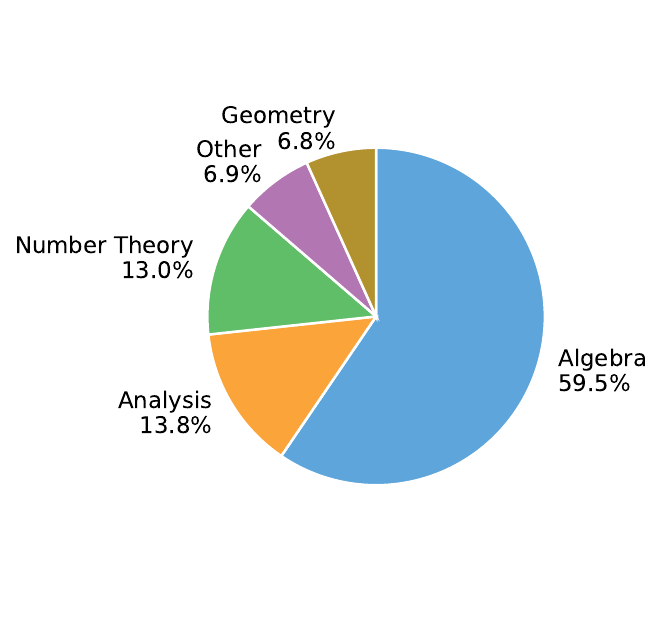}
\caption{Mathematical domain.}                                                                          
\label{fig:domain_pie}
\end{subfigure}\hfill                                                                                     
\begin{subfigure}[b]{0.33\linewidth}                                                                      
\centering
\includegraphics[width=0.9\linewidth]{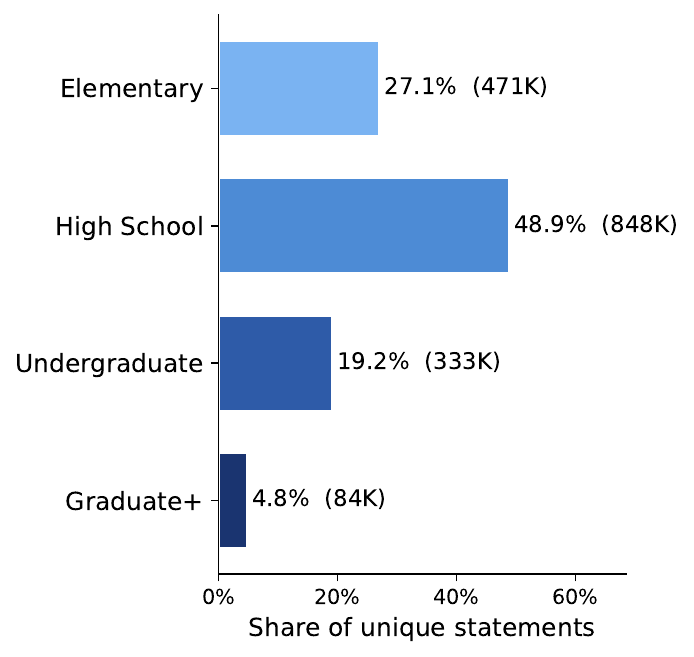}
\caption{Difficulty level.}                                                                             
\label{fig:difficulty_bar}
\end{subfigure}                                                                                           
\caption{OProofs Lean~4 corpus statistics.                                  
(a) Headline counts of statements, proofs, retrieval/feedback signals,
agentic trajectories, and training tokens.                                                              
(b) Mathematical domain distribution.                             
(c) Difficulty level distribution.}                                        
\label{fig:corpus_overview}
\end{figure*}

\subsection{Corpus Overview}

OProofs is a large-scale Lean~4 corpus constructed from both public formal resources and model-generated data. Public resources contribute formal statements and theorem--proof pairs, while our proof synthesis and agentic proving pipelines further expand the corpus with additional compiler-verified proofs and serialized proving trajectories. These trajectories are further decomposed into round-level repair examples for training. As a result, OProofs supports not only compiler-verified proof supervision, but also retrieval-grounded and feedback-conditioned agentic theorem proving.


Figure~\ref{fig:corpus_overview}(a) summarizes the current Lean~4 component of OProofs. The present version contains approximately 1.77M unique Lean statements paired with 6.86M compiler-verified proofs, together with 1.07M agentic proving trajectories. Beyond raw verified proofs, OProofs explicitly preserves the supervision signals required for retrieval-grounded and feedback-conditioned proof revision: 4.33M proofs carry retrieved proof context, 869k carry non-trivial compiler feedback from prior failed attempts, and 280k round-level repair examples are derived from the 164k trajectories that required at least one repair step. The frozen Lean corpus snapshot used in our current training pipeline comprises
51.8B tokens (prompt and chain-of-thought combined, tokenized with the Qwen3 tokenizer).

The corpus spans a broad spectrum of mathematics (Figure~\ref{fig:corpus_overview}(b)): Algebra (59.5\%), Analysis (13.8\%), Number Theory (13.0\%), and Geometry (6.8\%) dominate, with the remaining 6.9\% grouped as ``Other'' (covering Combinatorics, Logic, Probability, Topology, and Computation). Difficulty levels (Figure~\ref{fig:corpus_overview}(c)) range from Elementary arithmetic (27.1\%) through High School (48.9\%) and Undergraduate (19.2\%) to Graduate-level problems (4.8\%); domain and difficulty are inferred per statement by an LLM classifier (see Appendix~\ref{app:dd_prompt} for the prompt and validation protocol).
During iterative training, the underlying OProofs corpus continues to expand through newly verified proofs and repair data collected from the evolving prover. We emphasize that the statistics reported in Figure~\ref{fig:corpus_overview} cover only the Lean~4 theorem-proving component of OProofs, which is also the component we primarily analyze and release in this work. Auxiliary code, mathematical reasoning, and long-context reasoning corpora used during continued pretraining are excluded from these counts.

Table~\ref{tab:oproofs_comparison} compares OProofs with representative open Lean resources along supervision dimensions relevant to agentic formal theorem proving, including formal statements, compiler-verified proofs, natural-language rationales, compiler feedback, retrieved proof context, and multi-round repair signals. Among existing public resources, Nemotron-Math-Proofs is one of the closest to OProofs, as it already provides Lean~4 formal statements, compiler-verified proofs, natural-language reasoning traces, and compiler feedback. However, it does not explicitly release retrieved proof context or multi-round repair supervision. More broadly, most public resources expose only a subset of the signals needed for agentic formal theorem proving. OProofs is designed to close this gap by jointly providing retrieved proof context, compiler feedback, and repair supervision within a unified corpus for retrieval-grounded, feedback-conditioned proof revision.

\begin{table*}[t]
\centering
\caption{Comparison of OProofs with representative open Lean resources.}
\label{tab:oproofs_comparison}
\footnotesize
\setlength{\tabcolsep}{3pt}
\begin{tabular}{@{}lcccccc@{}}
\toprule
\textbf{Resource}
& \makecell[c]{\textbf{Formal}\\\textbf{stmt.}}
& \makecell[c]{\textbf{Verified}\\\textbf{proof}}
& \makecell[c]{\textbf{NL rationale}\\\textbf{/ CoT}}
& \makecell[c]{\textbf{Compiler}\\\textbf{feedback}}
& \makecell[c]{\textbf{Retrieved}\\\textbf{proof}}
& \makecell[c]{\textbf{Multi-round}\\\textbf{repair}} \\
\midrule
NuminaMath-LEAN
& \textcolor{green}{\ding{51}} & \textcolor{red}{\ding{55}} & \textcolor{red}{\ding{55}} & \textcolor{red}{\ding{55}} & \textcolor{red}{\ding{55}} & \textcolor{red}{\ding{55}} \\
Leanabell-SFT
& \textcolor{green}{\ding{51}} & \textcolor{green}{\ding{51}} & \textcolor{green}{\ding{51}} & \textcolor{red}{\ding{55}} & \textcolor{red}{\ding{55}} & \textcolor{red}{\ding{55}} \\
Lean-Workbook
& \textcolor{green}{\ding{51}} & \textcolor{red}{\ding{55}} & \textcolor{red}{\ding{55}} & \textcolor{red}{\ding{55}} & \textcolor{red}{\ding{55}} & \textcolor{red}{\ding{55}} \\
Leanabell-FormalStmt
& \textcolor{green}{\ding{51}} & \textcolor{red}{\ding{55}} & \textcolor{red}{\ding{55}} & \textcolor{red}{\ding{55}} & \textcolor{red}{\ding{55}} & \textcolor{red}{\ding{55}} \\
Workbook-proofs
& \textcolor{green}{\ding{51}} & \textcolor{green}{\ding{51}} & \textcolor{red}{\ding{55}} & \textcolor{red}{\ding{55}} & \textcolor{red}{\ding{55}} & \textcolor{red}{\ding{55}} \\
Goedel-Pset
& \textcolor{green}{\ding{51}} & \textcolor{red}{\ding{55}} & \textcolor{red}{\ding{55}} & \textcolor{red}{\ding{55}} & \textcolor{red}{\ding{55}} & \textcolor{red}{\ding{55}} \\
FineLeanCorpus
& \textcolor{green}{\ding{51}} & \textcolor{red}{\ding{55}} & \textcolor{red}{\ding{55}} & \textcolor{red}{\ding{55}} & \textcolor{red}{\ding{55}} & \textcolor{red}{\ding{55}} \\
FormalMATH-All
& \textcolor{green}{\ding{51}} & \textcolor{red}{\ding{55}} & \textcolor{red}{\ding{55}} & \textcolor{green}{\ding{51}} & \textcolor{red}{\ding{55}} & \textcolor{red}{\ding{55}} \\
Nemotron-Math-Proofs
& \textcolor{green}{\ding{51}} & \textcolor{green}{\ding{51}} & \textcolor{green}{\ding{51}} & \textcolor{green}{\ding{51}} & \textcolor{red}{\ding{55}} & \textcolor{red}{\ding{55}} \\
\midrule
\textbf{OProofs}
& \textcolor{green}{\ding{51}} & \textcolor{green}{\ding{51}} & \textcolor{green}{\ding{51}} & \textcolor{green}{\ding{51}} & \textcolor{green}{\ding{51}} & \textcolor{green}{\ding{51}} \\
\bottomrule
\end{tabular}
\end{table*}

\section{Experiments}\label{sec:experiments}

\paragraph{Benchmarks.}
We evaluate OProver on five Lean~4 theorem-proving benchmarks of varying difficulty and mathematical scope: MiniF2F~\citep{zheng2021minif2f} (244 high-school olympiad problems), MathOlympiadBench~\citep{lin2025goedelproverv2} (360 recent competition problems), ProofNet~\citep{azerbayev2023proofnet} (186 undergraduate-level theorems from textbook formalizations), ProverBench~\citep{ren2025deepseek} (325 problems spanning olympiad and undergraduate mathematics), and PutnamBench~\citep{tsoukalas2024putnambench} (672 Putnam competition problems, the hardest in the suite). This is also the standard evaluation suite used by recent open-source provers including Goedel-Prover-V2~\citep{lin2025goedelproverv2} and LongCat-Flash-Prover~\citep{wang2026longcat}, allowing direct comparison.

\paragraph{Baselines.}
We compare OProver against open-weight reasoning models (DeepSeek-V3.2 \citep{liu2025deepseek} and Kimi-K2.5 \citep{team2026kimi}) and open-weight whole-proof formal theorem provers, including Kimina-Prover \citep{wang2025kimina}, DeepSeek-Prover-V2 \citep{ren2025deepseek}, Leanabell-Prover-V2 \citep{ji2025leanabell}, Goedel-Prover-V2 \citep{lin2025goedelproverv2}, and LongCat-Flash-Prover \citep{wang2026longcat}. Our primary comparisons focus on Goedel-Prover-V2 (closest model scale at 32B) and LongCat-Flash-Prover (state-of-the-art open-weight whole-proof prover at 560B MoE / 27B active). For baseline results not re-evaluated by us, we mark the numbers with $\dagger$.

\paragraph{Evaluation protocol.}
We report Pass@$k$ as the primary metric, computed using the standard unbiased estimator
\[
\mathrm{Pass@}k = 1 - \frac{\binom{n-m}{k}}{\binom{n}{k}},
\]
where $n$ is the number of independent samples per statement and $m$ is the number of successful samples; specific values of $n$ and $k$ are reported in each table or caption. Unless otherwise noted, we report Pass@32 with $n=64$. OProver is evaluated under the agentic proving setup defined in \S\ref{sec:oprover_framework}, where one sample is a complete multi-round rollout (up to $R$ refinement rounds), and a sample is counted as successful if any of its attempts is verified by Lean~4. Baseline models are evaluated under their own protocols (whole-proof sampling, best-of-N with verifier filtering, or
whole-proof with TIR) to match each method's intended deployment. Section~\ref{sec:tts} provides a budget-controlled comparison that accounts for OProver's multi-round compute.

\subsection{Main Results}
\begin{table*}[t]
\centering
\caption{Theorem-proving performance (Pass@32, \%) of open-weight reasoning models and
prover models across multiple benchmarks. Best results are in bold and second-best
results are underlined. $\dagger$ marks scores not re-evaluated by us under our protocol.}
\label{tab:main_results}
\footnotesize
\setlength{\tabcolsep}{3pt}
\begin{tabular}{@{}lcccccc@{}}
\toprule
\textbf{Model}
& \textbf{\#Params}
& \makecell[c]{\textbf{MathOlympiad}\\\textbf{(Pass@32)}}
& \makecell[c]{\textbf{MiniF2F}\\\textbf{(Pass@32)}}
& \makecell[c]{\textbf{ProofNet}\\\textbf{(Pass@32)}}
& \makecell[c]{\textbf{ProverBench}\\\textbf{(Pass@32)}}
& \makecell[c]{\textbf{PutnamBench}\\\textbf{(Pass@32)}} \\
\midrule

\rowcolor{blue!8}
\multicolumn{7}{@{}c@{}}{\textit{Open-Weights Reasoning Models}} \\
DeepSeek-V3.2              & 671B & 14.7$\dagger$          & 77.9$\dagger$          & 20.4$\dagger$          & 42.8$\dagger$          & 5.8$\dagger$ \\
Kimi-K2.5                  & 1T   & 7.5$\dagger$           & 76.6$\dagger$          & 19.9$\dagger$          & 44.3$\dagger$          & 1.2$\dagger$ \\
\midrule

\rowcolor{blue!8}
\multicolumn{7}{@{}c@{}}{\textit{Open-Weights Prover Models}} \\
Kimina-Prover-8B           & 8B   & 7.6                    & 77.2                   & 13.0                   & 40.5                   & 2.3 \\
Kimina-Prover-72B          & 72B  & 11.5                   & 83.4                   & 16.3                   & 45.0                   & 4.0 \\
DeepSeek-Prover-V2-7B      & 7B   & 7.9                    & 73.9                   & 18.1                   & 45.7                   & 1.5 \\
DeepSeek-Prover-V2-671B    & 671B & 13.9$\dagger$          & 82.4$\dagger$          & 30.5$\dagger$          & 52.9$\dagger$          & 3.3$\dagger$ \\
Leanabell-Prover-V2-KM     & 8B   & --                     & 68.4$\dagger$          & 13.4$\dagger$          & 39.8$\dagger$          & -- \\
Leanabell-Prover-V2-DS     & 7B   & --                     & 76.6$\dagger$          & 23.7$\dagger$          & 47.8$\dagger$          & -- \\
Goedel-Prover-V2-8B        & 8B   & 9.9                    & 79.4                   & 17.3                   & 48.6                   & 0.7 \\
\quad w/ self-correction   & 8B   & --                     & 86.7$\dagger$          & --                     & --                     & -- \\
Goedel-Prover-V2-32B       & 32B  & 16.0                   & 85.8                   & 22.0                   & 51.0                   & 5.0 \\
\quad w/ self-correction   & 32B  & 20.3$\dagger$          & 90.4$\dagger$          & --                     & --                     & 8.6$\dagger$ \\
LongCat-Flash-Prover       & 560B & 16.9$\dagger$          & 84.4$\dagger$          & 19.9$\dagger$          & 49.9$\dagger$          & 4.9$\dagger$ \\
\quad whole-proof w/ TIR   & 560B & \textbf{27.5$\dagger$} & 90.2$\dagger$          & \textbf{36.1$\dagger$} & \underline{57.9$\dagger$} & \underline{10.4$\dagger$} \\
\midrule

\rowcolor{blue!8}
\multicolumn{7}{@{}c@{}}{\textit{Ours}} \\
OProver-8B                 & 8B   & 21.7                   & \underline{91.8}       & 31.9                   & 56.0                   & 9.0 \\
OProver-32B                & 32B  & \underline{22.8}       & \textbf{93.3}          & \underline{33.2}       & \textbf{58.2}          & \textbf{11.3} \\

\bottomrule
\end{tabular}
\end{table*}

Table~\ref{tab:main_results} compares OProver with representative open-weight reasoning models and open-weight prover models on five theorem-proving benchmarks. OProver-32B attains the best Pass@32 on three benchmarks (MiniF2F 93.3, ProverBench 58.2, PutnamBench 11.3) and the second-best on the remaining two (MathOlympiad 22.8, ProofNet 33.2), behind only LongCat-Flash-Prover with whole-proof TIR. This gives OProver-32B more top placements (three best, two second-best) than any other model in the table. Even OProver-8B outperforms Goedel-Prover-V2-32B on all five benchmarks despite having four times fewer parameters.

These results are notable because OProver-32B is a 32B dense model, whereas its closest competitor, LongCat-Flash-Prover, is a 560B Mixture-of-Experts model with roughly 27B active parameters per token---over 17$\times$ more total parameters. OProver-32B also outperforms the 671B DeepSeek-Prover-V2 on all five benchmarks, and improves over Goedel-Prover-V2-32B by 6.3 to 11.2 points per benchmark. Taken together, these comparisons indicate that the gains are not a consequence of model scale, but are driven by the combination of retrieval, multi-turn compiler feedback, and iterative test-time refinement.

The gains span benchmarks of varying difficulty: from competition-style problems on MiniF2F, through textbook-level formalizations on ProverBench, to the substantially harder Putnam-style problems on PutnamBench, where absolute success rates remain below 12\% for all methods. Achieving the best results across this entire spectrum suggests that OProver's improvements reflect a broadly effective proving strategy rather than specialization on any single benchmark family.

\subsection{Test-Time Scaling}
\label{sec:tts}
We next study whether OProver can effectively benefit from additional test-time compute. We define
\[
\mathrm{BestPass}(B) = \max_{R \cdot k = B} \mathrm{Pass}(R, k),
\]
where $R$ is the number of refinement rounds, $k$ is the number of samples per round, and $B$ is the total test-time budget. This metric measures the best success rate achievable under a fixed total budget, regardless of how the budget is allocated between iterative refinement and independent sampling.

\begin{figure*}[t]
\centering
\includegraphics[width=\linewidth]{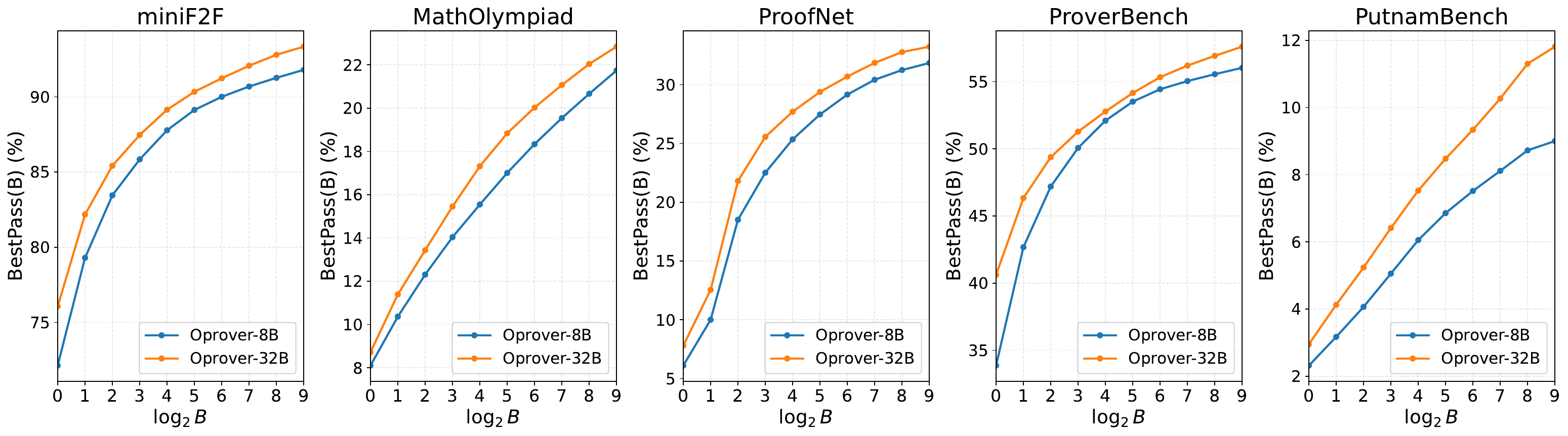}
\caption{Test-time scaling of OProver under a fixed total budget $B$. For each benchmark, we report $\mathrm{BestPass}(B)$ (defined in \S\ref{sec:tts}), the best success rate achievable under budget $B$ across all $(R, k)$ allocations. Both OProver-8B and OProver-32B improve consistently with larger budgets, while the gains exhibit clear diminishing returns.}
\label{fig:bestpass_budget}
\end{figure*}

Figure~\ref{fig:bestpass_budget} shows that both OProver-8B and OProver-32B improve consistently as the total budget increases on all five benchmarks. For example, when increasing the budget from $B=8$ to $B=256$, OProver-32B improves from 87.5 to 92.8 on MiniF2F, from 15.5 to 22.0 on MathOlympiad, from 25.6 to 32.8 on ProofNet, from 51.3 to 56.9 on ProverBench, and from 6.4 to 11.3 on PutnamBench. Similar gains are observed for OProver-8B. These results indicate that test-time scaling is robust across both model sizes and benchmark types.

At the same time, the gains show clear diminishing returns, with the improvement per budget doubling shrinking steadily as $B$ grows. On ProofNet with OProver-32B, for instance, doubling $B$ from $8$ to $16$ yields a 2.1-point gain (25.6 $\rightarrow$ 27.7), but the same doubling from $128$ to $256$ adds only 0.9 points (31.9 $\rightarrow$ 32.8). The same pattern is visible on every benchmark and for both model sizes. The rate of diminishing returns is also benchmark-dependent: on MiniF2F and ProverBench, where baseline success rates are already high, $\mathrm{BestPass}$ approaches saturation more quickly, whereas on PutnamBench the gains remain comparatively steady throughout the budget range, consistent with the much lower baseline leaving more room for additional compute to help.

To further study how test-time compute should be allocated, we fix the total budget $B$ and plot $\mathrm{Pass}(R, B/R)$ as a function of the refinement depth $R$, which isolates the trade-off between deeper multi-turn interaction and wider independent sampling under the same total compute budget.

\begin{figure*}[t]
\centering
\includegraphics[width=\textwidth]{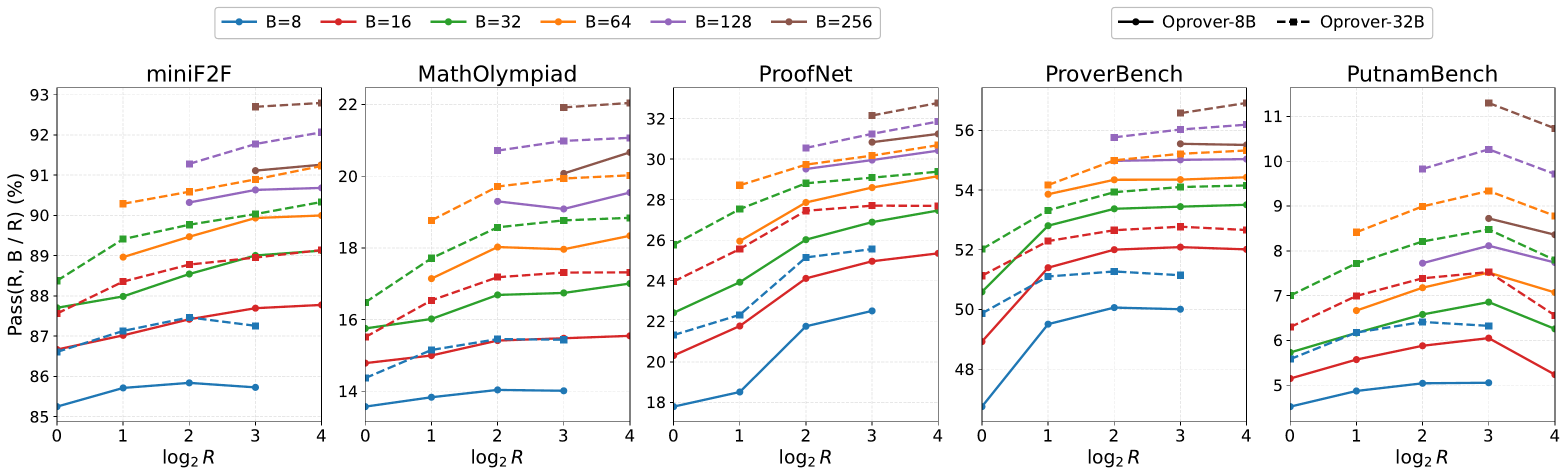}
\caption{Performance under fixed total budget $B$ as a function of refinement depth $R$. Each curve plots $\mathrm{Pass}(R, B/R)$, isolating the trade-off between deeper interaction and wider sampling. On most benchmarks performance keeps improving as $R$ grows up to $R=16$, while on the hardest benchmark (PutnamBench) the optimum settles around $R=8$, indicating that very narrow sampling can hurt when single-chain success probability is low.}
\label{fig:fixed_budget_tradeoff}
\end{figure*}

Figure~\ref{fig:fixed_budget_tradeoff} reveals that the optimal allocation between refinement depth $R$ and sampling width $k=B/R$ is governed by the per-chain success probability of the benchmark. Increasing $R$ lets each chain exploit Lean feedback to repair its own proof, while increasing $k$ spreads the budget across more independent attempts; under a fixed budget these two forces compete.

On MiniF2F, MathOlympiad, ProofNet, and ProverBench, single-chain success probabilities are already substantial, so the marginal gain from additional refinement rounds remains positive throughout. Performance improves monotonically with $R$, and the best configuration under any medium-to-large budget is consistently attained at $R=16$ for both model sizes. $1-(1-p)^k$ describes the probability that at least one of $k$ independent chains succeeds. When $p$ is small---as on PutnamBench, where per-chain success stays in the $5$--$11\%$ range---this probability is highly sensitive to $k$: halving $k$ to double $R$ costs more in lost exploration than it gains in deeper refinement. Empirically, the optimum settles at $R=8$ rather than $R=16$ for every $B\geq 16$ on both model sizes; at $B=256$, OProver-32B reaches $11.3\%$ at $R=8$ but drops to $10.7\%$ at $R=16$. The same effect appears on easier benchmarks under small budgets: when $B=8$, pushing $R$ to $16$ forces $k=1$, eliminating all parallel exploration, and the optimum shifts to $R=4$--$8$.

In summary, the optimal allocation is benchmark-dependent. On easier-to-moderate benchmarks, $R=16$ is consistently optimal once $B\geq 16$. On the hardest benchmark, the optimum saturates at $R=8$---the point where additional refinement depth no longer compensates for the reduced sampling width.

\subsection{Effectiveness of Iterative Post-Training}
\label{sec:iteration_progression}

\begin{wrapfigure}{r}{0.45\linewidth}
\vspace{-0.6em}
\centering
\includegraphics[width=\linewidth]{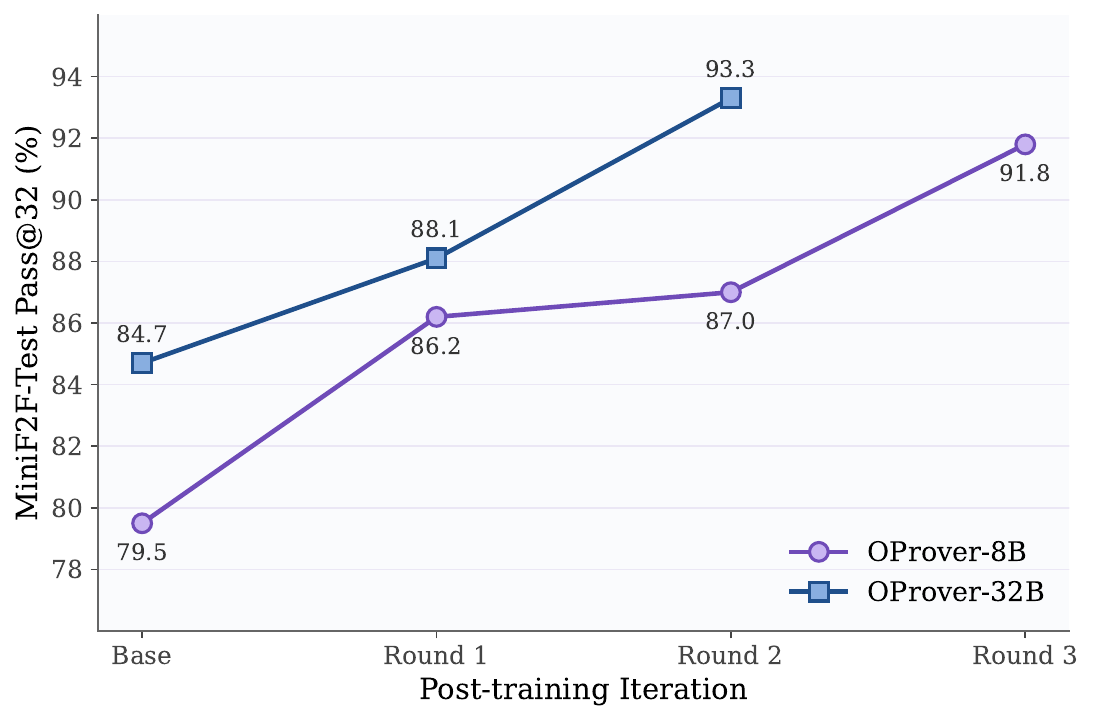}
\vspace{-1.4em}
\caption{MiniF2F-Test Pass@32 across post-training iterations. Both OProver-8B and OProver-32B improve monotonically.}
\label{fig:iteration_progression}
\vspace{-0.8em}
\end{wrapfigure}


To examine whether the iterative co-evolution loop between OProver and OProofs (\S\ref{sec:training}) yields measurable improvements at each step, we evaluate Pass@32 on MiniF2F-Test for every intermediate checkpoint. Figure~\ref{fig:iteration_progression} shows that both model sizes improve monotonically across iterations. OProver-8B rises from 79.5 (Base, after continued pretraining only) to 86.2, 87.0, and 91.8 after Rounds~1, 2, and 3, for a total gain of 12.3 points. OProver-32B rises from 84.7 to 88.1 and 93.3 after Rounds~1 and 2, for a total gain of 8.6 points. These gains confirm that recirculating verified proofs and repair trajectories from the current prover into OProofs (and into the retrieval memory) produces a stronger prover in the subsequent iteration, rather than saturating after the first post-training round.

\subsection{Ablation Studies}

\begin{table*}[t]
\centering
\caption{Ablation on multi-turn compiler feedback and retrieval augmentation. ``-FB'' removes multi-turn compiler feedback and keeps only the first interaction round (retrieval retained). ``-FB, -RAG'' additionally removes retrieved
proofs from the prompt. All numbers are Pass@32 (\%).}
\label{tab:ablation_feedback}
\scriptsize
\setlength{\tabcolsep}{4pt}
\renewcommand{\arraystretch}{1.05}
\begin{tabular}{@{}llccccc@{}}
\toprule
\textbf{Model} & \textbf{Variant}
& \makecell[c]{\textbf{MathOlympiad}}
& \makecell[c]{\textbf{MiniF2F}}
& \makecell[c]{\textbf{ProofNet}}
& \makecell[c]{\textbf{ProverBench}}
& \makecell[c]{\textbf{PutnamBench}} \\
\midrule
OProver-8B  & Full        & 21.7 & 91.8 & 31.9 & 56.0 & 9.0  \\
OProver-8B  & -FB         & 15.8 & 87.7 & 22.4 & 50.6 & 5.7  \\
OProver-8B  & -FB, -RAG   & 12.4 & 86.2 & 22.4 & 49.9 & 4.1  \\
\midrule
OProver-32B & Full        & 22.8 & 93.3 & 33.2 & 58.2 & 11.3 \\
OProver-32B & -FB         & 16.5 & 88.4 & 25.8 & 52.0 & 7.0  \\
OProver-32B & -FB, -RAG   & 14.8 & 87.9 & 24.7 & 51.1 & 5.9  \\
\bottomrule
\end{tabular}
\end{table*}

We ablate two core components of OProver: multi-turn compiler feedback and retrieval augmentation. Specifically, \textit{w/o feedback} keeps only the first interaction round, i.e., the prover generates 32 candidates in round 1 without subsequent refinement (retrieval retained), while \textit{w/o feedback, w/o RAG} additionally removes retrieved proofs from the prompt.

Table~\ref{tab:ablation_feedback} shows that both components contribute to OProver's performance, but with markedly different magnitudes. Removing compiler feedback causes the largest and most consistent degradation across benchmarks and model sizes, confirming that iterative refinement is the primary driver of OProver's gains. For OProver-8B, Pass@32 drops from 21.7 to 15.8 on MathOlympiad, from 91.8 to 87.7 on MiniF2F, from 31.9 to 22.4 on ProofNet, from 56.0 to 50.6 on ProverBench, and from 9.0 to 5.7 on PutnamBench. For OProver-32B, the corresponding drops are from 22.8 to 16.5, from 93.3 to 88.4, from 33.2 to 25.8, from 58.2 to 52.0, and from 11.3 to 7.0.

Further removing RAG leads to a smaller but consistent degradation, indicating that retrieval provides complementary benefits beyond feedback alone. For OProver-32B, the additional drops range from 0.5 points on MiniF2F (88.4 $\rightarrow$ 87.9) to 1.7 points on MathOlympiad (16.5 $\rightarrow$ 14.8), with intermediate drops on ProofNet (1.1), ProverBench (0.9), and PutnamBench (1.1). While feedback mainly helps repair partially correct proofs over multiple rounds, retrieval improves proof grounding by exposing relevant premises and lemmas before refinement begins. The combined results therefore suggest that OProver's gains come from the synergy between retrieval-augmented grounding and compiler-guided iterative refinement, rather than from single-round sampling alone, with feedback contributing the dominant share.

\section{Related Work}

\subsection{Proof Generation and Search}

Neural theorem proving with language models has developed along two closely related directions: proof generation and guided search. Early work such as GPT-f \citep{polu2020generative} and Proof Artifact Co-Training (PACT) \citep{han2021proof} showed that formal proofs can be modeled as generated sequences and that proof artifacts provide useful supervision for neural provers. In parallel, search-based approaches such as HyperTree Proof Search (HTPS) \citep{lample2022hypertree} demonstrated that neural guidance can be integrated with symbolic exploration to improve proof discovery.

Within Lean theorem proving, a major line of work studies tactic-level interaction and state-space search. LeanDojo and ReProver \citep{yang2023leandojo} introduced a retrieval-augmented interactive environment and highlighted premise selection as a core challenge for large-library theorem proving. Subsequent systems strengthened this paradigm with learned critics, planners, and feedback-aware search. InternLM2.5-StepProver \citep{wu2025internlm2} scales expert iteration with critic-guided proving, BFS-Prover \citep{xin2025bfs} revisits best-first search with preference optimization from compiler feedback, Bourbaki \citep{zimmer2025bourbaki} formulates theorem proving as a goal-conditioned decision process with structured subgoal search, and DeepSeek-Prover-V1.5 \citep{xin2024deepseek} combines proof-assistant feedback with search-time exploration through an MCTS-style procedure.

A second line increasingly treats theorem proving as whole-proof generation. DeepSeek-Prover-V1.5~\citep{xin2024deepseek} demonstrated that scaling synthetic theorem--proof data, combined with proof-assistant feedback, substantially improves end-to-end Lean proof generation. More recent systems extend this paradigm with stronger base models, longer-horizon reasoning, subgoal decomposition, and reinforcement learning, including Goedel-Prover~\citep{lin2025goedel}, DeepSeek-Prover-V2~\citep{ren2025deepseek}, Kimina-Prover~\citep{wang2025kimina}, and Seed-Prover~\citep{chen2025seed}. Together, these results establish whole-proof generation as a competitive paradigm for modern formal theorem proving.

\subsection{Verifier-Guided Refinement and Agentic Proving}

A growing body of work argues that formal theorem proving should not be treated as a single-pass generation problem, since proof assistants provide dense and semantically meaningful feedback on failed attempts. Baldur \citep{first2023baldur} is an influential early example in this direction: it combines whole-proof generation with a dedicated repair model, showing that unsuccessful proofs can be revised rather than discarded. In Lean, Lean-STaR \citep{lin2024lean} interleaves informal reasoning with formal proving steps, while DeepSeek-Prover-V1.5 \citep{xin2024deepseek} incorporates proof-assistant feedback into both reinforcement learning and search.

Recent work has made this interaction pattern increasingly explicit and increasingly agentic. Leanabell-Prover-V2 \citep{ji2025leanabell} studies verifier-integrated reasoning with reinforcement learning, StepFun-Prover \citep{shang2025stepfun} trains tool-integrated reasoning models that refine proofs using environment feedback, and Prover Agent \citep{baba2025prover} coordinates informal reasoning, formal proving, and auxiliary lemma generation within an agent-style architecture. \citet{gupta2025tool} further show that multi-turn proof repair can be learned directly from interactive trajectories rather than being introduced only at inference time. More recent systems strengthen this trend: Seed-Prover \citep{chen2025seed} combines Lean feedback with proved lemmas and self-summarization to support iterative whole-proof refinement, and LongCat-Flash-Prover \citep{wang2026longcat} advances agentic tool-integrated reinforcement learning for formal reasoning. Collectively, these works show that verifier feedback is not merely a binary correctness signal, but a rich supervision source for iterative and long-horizon proof refinement.

\subsection{Formal Data Synthesis and Process Supervision}

Recent progress in formal theorem proving has also depended critically on advances in data construction. Because high-quality formal theorem--proof pairs remain scarce relative to the scale required for modern model training, many recent systems rely on synthetic proof generation, large-scale formalization, and iterative data expansion. Lean Workbook \citep{ying2024lean} contributes a large-scale collection of Lean problems formalized from natural-language mathematics, while TheoremForge \citep{tao2026theoremforge} studies how agentic workflows can be used to synthesize formal reasoning data under limited annotation budgets. Large-scale prover pipelines such as DeepSeek-Prover-V2~\citep{ren2025deepseek}, Goedel-Prover~\citep{lin2025goedel}, and Goedel-Prover-V2~\citep{lin2025goedelproverv2} further demonstrate that synthetic data construction is now central to scaling formal reasoning systems.

A related line of work investigates how supervision can be enriched beyond final verified proofs. Draft, Sketch, and Prove \citep{jiang2022draft} shows that informal proof sketches can guide formal proof generation through intermediate reasoning structure, and ProofNet \citep{azerbayev2023proofnet} establishes a benchmark connecting undergraduate-level mathematical statements, informal proofs, and formal verification. More recent efforts further incorporate critic signals and tool feedback into the supervision interface. CriticLean \citep{peng2025criticlean} introduces critic-guided training for semantic fidelity in Lean formalization, Autoformalizer with Tool Feedback \citep{guo2025autoformalizer} uses compiler feedback and consistency signals to iteratively improve generated formal statements, and STP \citep{dong2025stp} moves toward self-improving theorem proving through iterative cycles of conjecturing and proving.

Our work is situated at the intersection of these directions. OProofs extends formal proof corpora toward trajectory-level supervision by preserving multi-turn interaction histories, including failed attempts, retrieved proof context, compiler feedback, and subsequent repairs. The OProver post-training loop further couples data construction with policy improvement by adding newly verified proofs to OProofs and indexing them into retrieval memory, while using repair examples and harder unresolved cases for supervised and reinforcement learning. In this sense, our work unifies formal data synthesis and process supervision within a single training framework for agentic formal theorem proving. Compared to prior self-improving systems such as STP~\citep{dong2025stp} and Seed-Prover~\citep{chen2025seed}, which iterate on statement--proof pairs, OProver explicitly preserves and learns from serialized multi-turn proving trajectories, enabling retrieval-grounded and feedback-conditioned proof revision as a learned policy rather than an inference-time wrapper.

\section{Conclusion}
We present OProver, a unified framework for agentic formal theorem proving in Lean~4. OProver treats proving as a multi-round refinement loop in which a trained policy revises failed proof attempts using retrieved compiler-verified proofs and Lean~4 compiler feedback, unifying retrieval, feedback, and iterative repair within a single proving policy. To support this framework, we construct OProofs, a large-scale Lean~4 corpus containing approximately 1.77M statements, 6.86M compiler-verified proofs, and serialized agentic proving trajectories that record retrieval contexts, failed attempts, compiler feedback, and subsequent repairs. OProofs provides Lean-specific supervision across continued pretraining, supervised fine-tuning, and reinforcement learning, and serves as the memory indexed for retrieval at proving time. Using OProofs, we train OProver-8B and OProver-32B through continued pretraining followed by iterative post-training with SFT and RL, where verified proofs and repair trajectories produced by the current prover are folded back into the corpus for subsequent iterations. Across five Lean~4 benchmarks, OProver-32B attains the best Pass@32 on MiniF2F, ProverBench, and PutnamBench, and the second-best on MathOlympiad and ProofNet, achieving more top placements than any prior open-weight whole-proof prover.

\clearpage

\section{Contributions and Acknowledgments}

Multimodal Art Projection (M-A-P) is a non-profit open-source AI research community, run by donations.
The community members are working on research topics in a wide range of spectrum, including but not limited to the pre-training paradigm of foundation models, large-scale data collection and processing, and the derived applications on coding, reasoning, and music generation.

\textbf{Core Contributors}
\begin{multicols}{2}
    \begin{itemize}
        \item David Ma, M-A-P
        \item Kaijing Ma, M-A-P
    \end{itemize}
\end{multicols}

\textbf{Contributors}
\begin{multicols}{2}
    \begin{itemize}
        \item Shawn Guo, M-A-P
        \item Enduo Zhao, M-A-P, Monolith
        \item Yunfeng Shi, M-A-P
        \item Jiajun Shi, M-A-P, Beihang University
        \item Jiameng Huang, M-A-P, Peking University
        \item Bingrui Li, M-A-P, Peking University
        \item Zhenzhu Yang, M-A-P
        \item Zhaoxiang Zhang, CASIA
    \end{itemize}
\end{multicols}

\textbf{Corresponding Authors}
\begin{multicols}{2}
    \begin{itemize}
        \item Gavin Cheung, M-A-P
        \item Jiaheng Liu, M-A-P, Nanjing University
        \item Zili Wang, M-A-P
    \end{itemize}
\end{multicols}

\newpage

\bibliography{main.bib}

\newpage
\appendix

\section{Additional Method Details}
\label{app:method_details}

This appendix provides additional details on retriever selection, repair instance construction, data recirculation, training and evaluation hyperparameters, and pretraining data recall.

\subsection{Retriever Selection}
\label{app:retrieval_memory}

Section~\ref{sec:oprover_framework} describes how OProver retrieves top-$k$ compiler-verified proofs from a memory $\mathcal{M}$ at each proving round. This appendix details the retriever model selection.

We compared two candidate retrievers: Qwen3-8B-Embedding and Goedel-Prover-V2-8B. The comparison was conducted on a held-out retrieval set of 1{,}000 query statements and 9{,}000 candidate retrieval documents.

\paragraph{Similarity score distribution.}
Qwen3-8B-Embedding produced similarity scores spanning approximately $[0.20, 0.99]$, whereas Goedel-Prover-V2-8B produced scores concentrated in a narrow high-similarity range $[0.85, 0.99]$. The broader spread of Qwen3-8B-Embedding provides a more discriminative ranking signal: when score variance is small across candidates, top-$k$ selection becomes effectively random, whereas a broader spread aligns top-$k$ selection with semantic relevance.

\paragraph{Pairwise LLM-as-a-judge evaluation.}
We further evaluated retrieval quality using GPT-4o in a pairwise LLM-as-a-judge setup. For each query, GPT-4o compared the top-5 retrieved theorem contexts returned by the two retrievers. Over 1{,}000 query-level comparisons, Qwen3-8B-Embedding was preferred in 667 cases (66.7\%), while Goedel-Prover-V2-8B was preferred in 333 cases (Table~\ref{tab:retriever_comparison}). Based on these results, OProver uses Qwen3-8B-Embedding as the default retriever.

\begin{table}[t]
\centering
\small
\caption{Retriever comparison on 1{,}000 queries with 9{,}000 candidates. Relevance is assessed by GPT-4o
pairwise judgment over top-5 retrieved contexts.}
\label{tab:retriever_comparison}
\setlength{\tabcolsep}{4pt}
\begin{tabular}{lcc}
\toprule
\textbf{Retriever} & \textbf{Sim.\ spread} & \textbf{Judge pref.} \\
\midrule
Goedel-Prover-V2-8B & $[0.85, 0.99]$ & 333 / 1000 \\
Qwen3-8B-Embedding  & $[0.20, 0.99]$ & 667 / 1000 \\
\bottomrule
\end{tabular}
\end{table}

\subsection{Repair Instance Construction Details}
\label{app:repair_instance_construction}

Section~\ref{sec:oproofs} defines a round-level repair instance as $(s, \mathcal{R}, p^{-}, f^{-}, p^{+})$. We describe here how such instances are extracted from raw proving trajectories.

\paragraph{Extraction.}
Given an agentic proving trajectory of $T$ rounds $(p_1, f_1, p_2, f_2, \ldots, p_T)$, each transition $(p_{t-1}, f_{t-1}) \rightarrow p_t$ for $t \geq 2$ becomes one repair instance, paired with the retrieval context $\mathcal{R}_t$ used at round $t$. A trajectory of $T$ rounds therefore yields up to $T-1$ repair instances.

\paragraph{Filtering.}
We discard instances in which (i) $p_{t-1}$ is empty or syntactically malformed before verification, (ii) $f_{t-1}$ exceeds 8{,}000 tokens (these are dominated by repeated boilerplate errors), or (iii) $p_t$ differs from $p_{t-1}$ by fewer than 3 tokens (near-no-op revisions). The retained repair instances form the primary supervision substrate for feedback-aware post-training.

\paragraph{Deduplication.}
Repair instances are deduplicated by exact match on $(p^{-}, f^{-}, p^{+})$ to avoid multiple trajectories producing identical local transitions.

\subsection{Data Recirculation Details}
\label{app:data_recirculation}

Section~\ref{sec:training} describes how each post-training iteration recirculates the prover's rollouts into OProofs, the retrieval memory, the SFT repair set, and the RL hard-case set. We provide additional implementation details below.

\paragraph{Routing rules.}
After each iteration, each rollout is routed according to its outcome and the success rate of its theorem's group (the $n$ rollouts sampled for the same theorem):
\begin{itemize}
\item Rollouts that yield a Lean-verified proof contribute their final proof to $\mathcal{P}_t^{+}$, which is appended to OProofs $\mathcal{D}_{t+1}$ and re-indexed into the retrieval memory $\mathcal{M}_{t+1}$.
\item Repair instances extracted from successful rollouts are added to the SFT set $\mathcal{B}_t$.
\item Theorems whose group success rate is in $(0, 1)$---that is, not solved by all rollouts and not failed by all rollouts---are retained as hard cases $\mathcal{H}_t$ for the next round of RL.
\end{itemize}
The routing follows DAPO-style group filtering, which empirically reduces gradient variance by excluding both fully-solved and fully-failed groups.

\subsection{Training and Evaluation Hyperparameters}
\label{sec:exp_setup}

\paragraph{Continued pretraining.}
OProver-Base is obtained by continued pretraining on the 65B-token mixture described in Section~\ref{sec:training}, using AdamW with peak learning rate $5 \times 10^{-5}$, cosine schedule with 3\% warmup, global batch size 512, and sequence length 8192.

\paragraph{Supervised fine-tuning.}
OProver-8B and OProver-32B are fine-tuned with global sequence length 40{,}960 and global batch size 64. Learning rates are $2 \times 10^{-5}$ for OProver-8B and $5 \times 10^{-5}$ for OProver-32B. Cross-entropy loss is computed only on the target proof attempt $p_t$.

\paragraph{Reinforcement learning.}
We adopt GSPO with learning rate $2 \times 10^{-6}$, training batch size 256, PPO mini-batch size 256, and $n=8$ rollouts per theorem. The maximum prompt length is 14{,}000 tokens. The maximum response length is 24{,}000 tokens for OProver-8B and 20{,}000 tokens for OProver-32B. The maximum number of refinement rounds during RL is 4 for OProver-8B and 2 for OProver-32B. Rollout decoding uses temperature 1.0 and top-$p$ 0.999. The per-round verification reward is $r_t = 0.8 + 0.2 \cdot \mathbb{1}[\text{format correct}]$ when $\mathcal{V}(p_t) = $ verified, and $r_t = 0$ otherwise. Advantages are computed by group-relative normalization pooled across the $n \times R$ rounds for the same theorem. Lean verification during RL uses proof reconstruction with fully explicit proofs, a per-request timeout of 120s, and a heartbeat cap of 2M.

\paragraph{Evaluation.}
At evaluation time, we use the same multi-turn proving interface as RL training. Generation uses temperature 1.0, top-$p$ 0.999, and a maximum response length of 32{,}000 tokens. Proof verification uses the Kimina Lean server (Lean 4.15.0) with proof reconstruction and fully explicit proofs, a per-request timeout of 240s, and a heartbeat cap of 4M.

\subsection{Pretraining Data Recall}
\label{app:pretraining_data_recall}

We recall mathematics-adjacent examples from the pretraining corpus through a two-stage pipeline.

\paragraph{Stage 1: iterative fastText recall.}
We initialize a seed set with verified proofs and repair instances from OProofs, and train a fastText classifier to distinguish seed documents from background pretraining documents. The trained model is applied to the pretraining corpus to retrieve a high-recall candidate pool. High-scoring candidates are merged back into the seed set, after which the classifier is retrained for another retrieval round. This train--retrieve--expand cycle is repeated to broaden coverage beyond exact Lean syntax matches.

\paragraph{Stage 2: embedding-based precision filter.}
After the final fastText iteration, both seed examples and recalled candidates are encoded with a small embedding model trained on the same domain data. Candidates whose nearest-seed cosine similarity falls below a threshold $\tau$ are discarded. The retained documents form the recalled pretraining subset, which is mixed into the CPT data described in Section~\ref{sec:training}.
\section{Prompt Templates and Feedback Format}
\label{app:prompt_templates}

This appendix specifies the prompt template and the feedback serialization used by OProver during both training and inference, following the state formulation $X_t = (s, \mathcal{R}_t, p_{t-1}, f_{t-1})$ from Section~\ref{sec:oprover_framework}.

\definecolor{OProverBlueBg}{RGB}{245,249,253}
\definecolor{OProverBlueFrame}{RGB}{72,114,160}
\definecolor{OProverBlueTitle}{RGB}{72,114,160}

\tcbset{
oproverbox/.style={
enhanced,
breakable,
colback=OProverBlueBg,
colframe=OProverBlueFrame,
colbacktitle=OProverBlueTitle,
coltitle=white,
boxrule=0.7pt,
arc=1.2mm,
left=1mm,
right=1mm,
top=0.8mm,
bottom=0.8mm,
title filled,
fonttitle=\bfseries
}
}

\subsection{Prompt Template}
\label{app:prompt_format}

We use a single prompt template across all rounds. At round $t$, the model conditions on the target Lean~4 statement, the top-$k$ retrieved references, and the immediately preceding proof attempt with its Lean feedback. When $t=1$, the previous-attempt and feedback fields are left empty.

\begin{tcblisting}{oproverbox,
title=Unified Prompt Template at Round $t$,
listing only,
listing options={
basicstyle=\ttfamily\small,
breaklines=true,
breakatwhitespace=true,
columns=fullflexible,
keepspaces=true
}}
**Current Task:**
Complete the following Lean 4 code:
{FORMAL_STATEMENT}

Before producing the Lean 4 proof, first provide a concise proof plan
summarizing the intended strategy, key lemmas, and proof structure.
If a previous attempt and Lean feedback are provided, revise the proof
accordingly.

Reference theorems and proofs:
{RETRIEVED_PROOF_1}
{RETRIEVED_PROOF_2}

Previous Failed Attempt:
{PREVIOUS_PROOF_ATTEMPT}

Lean Feedback:
{LEAN_ERROR_MESSAGES}
\end{tcblisting}

\subsection{Feedback Serialization}
\label{app:feedback_format}

Lean feedback is serialized as the raw plain-text diagnostic returned by the Lean~4 compiler for the immediately preceding proof attempt, with no projection into a fixed error taxonomy. This preserves identifier names, expected/actual types, and goal contexts that targeted revision often depends on. An example diagnostic, after stripping ANSI control codes, looks like:

\begin{tcblisting}{oproverbox,
title=Example Lean Feedback,
listing only,
listing options={
basicstyle=\ttfamily\footnotesize,
breaklines=true,
breakatwhitespace=true,
columns=fullflexible,
keepspaces=true
}}
type mismatch
h0
has type
Nat.succ ?m.123
but is expected to have type
Real

tactic 'linarith' failed to find a contradiction
case h
x y : Real
h0 : x <= y
h1 : y <= 0
turnstile False
\end{tcblisting}

\subsection{Common Feedback Categories}
\label{app:feedback_categories}

Although feedback is passed in its raw textual form, the underlying diagnostics fall into a few recurring categories, which we summarize in Table~\ref{tab:feedback_types} together with the typical repair actions we observe across multi-round refinement.

\begin{table}[t]
\centering
\caption{Common Lean feedback categories observed in OProver's proving trajectories, with the typical repair actions taken in the subsequent round.}
\label{tab:feedback_types}
\small
\begin{tabular}{p{0.30\linewidth} p{0.62\linewidth}}
\toprule
\textbf{Feedback type} & \textbf{Typical repair action} \\
\midrule
Unsolved goals / goal mismatch &
Introduce missing intermediate lemmas or case splits, restate the target in a more convenient form, or align the local proof state before continuing. \\
Tactic or automation failure &
Replace brittle one-shot automation with explicit derivation steps, reorder tactics, or simplify the context before invoking automation such as \texttt{linarith}, \texttt{omega}, or \texttt{simp}. \\
Type or term mismatch &
Rewrite ill-typed expressions, add explicit coercions or binders, and ensure intermediate terms match the expected type. \\
Unknown identifier / namespace / notation error &
Qualify names explicitly, open the required namespace, fix notation, or replace unavailable lemmas with in-scope equivalents. \\
Incomplete induction or case split &
Add missing branches, make parity or sign cases explicit, and construct the required witnesses before closing the proof. \\
Timeout / heartbeat exhaustion &
Shorten proof scripts, reduce expensive search, and replace heavy automation with more structured local reasoning. \\
\bottomrule
\end{tabular}
\end{table}

In raw OProver logs, a non-verifier failure mode \texttt{no\_lean\_code\_found} also appears, reflecting generation or proof-extraction failure rather than a Lean compiler diagnostic. We exclude it from Table~\ref{tab:feedback_types} since it does not represent native verifier feedback.

\section{Effectiveness of Agentic Reasoning}
\label{app:agentic_reasoning}

This appendix provides qualitative evidence for the effectiveness of the agentic refinement loop in OProver. In our setting, the policy does not generate a proof in a single shot. Instead, it iteratively refines candidate proofs by conditioning on retrieved formal
references and Lean verifier feedback. The examples below illustrate how these two sources of information support grounded repair behavior across rounds. For space, all proof excerpts and error messages are abridged but preserve the essential failure mode and the
corresponding repair.

\subsection{Roles of Retrieval and Compiler Feedback}

Retrieved formal references are most useful in three recurring ways. First, they help ground lemma names, rewrite rules, and tactic idioms that are easy to miss in one-shot generation. Second, they provide short structural hints about how to decompose a goal into
subgoals that better match Lean's expected proof state. Third, they expose proof patterns that help the model avoid common formulation errors, such as missing coercions, inappropriate tactic ordering, or mismatched algebraic normal forms. In practice, these signals
are lightweight, but they often suffice to redirect a stalled proof attempt toward a more verifiable local proof structure.

Lean feedback provides a complementary signal: it identifies concrete verifier-side failures that can be mapped to local edits in the next round. Rather than projecting compiler outputs into a hand-designed taxonomy during inference, OProver conditions directly on
the original textual diagnostics. This preserves fine-grained information about the actual source of failure---including type mismatches, missing identifiers, unresolved goals, or tactic failures. Common categories of such feedback and the corresponding repair
actions are listed in Table~\ref{tab:feedback_types} of Appendix~\ref{app:prompt_templates}.

\paragraph{Presentation note.}
For compatibility with the \texttt{pdflatex} toolchain, the proof excerpts below normalize Lean Unicode notation into an ASCII display form while preserving the logical structure of each repair trace.

\subsection{Case Study A: Putnam 1963 A3}

This example illustrates a simple but representative local repair. The first-round attempt introduces an ill-typed contradiction block; the second-round repair removes the offending fragment and restores a compilable scaffold for subsequent refinement.

\begin{tcblisting}{oproverbox,
title=Round 1 Proof Attempt,
listing only,
listing options={
basicstyle=\ttfamily\footnotesize,
breaklines=true,
breakatwhitespace=true,
columns=fullflexible,
keepspaces=true}}
import Mathlib

set_option maxHeartbeats 2000000
open Nat Set Topology Filter

theorem putnam_1963_a3
(P : Nat -> (Real -> Real) -> (Real -> Real))
(hP : P 0 = id /\ forall i y, P (i + 1) y = P i (fun x => x * deriv y x - i * y x))
(n : Nat)
(hn : 0 < n)
(f y : Real -> Real)
(hf : ContinuousOn f (Ici 1))
(hy : ContDiffOn Real n y (Ici 1))
(hy1 : ContDiffAt Real n y 1) :
(forall i, i < n -> deriv^[i] y 1 = 0) /\ (Ici 1).EqOn (P n y) f <->
forall x, x >= 1 -> y x = integral ... := by
have h1 : False := by
have h2 : forall (alpha : Nat -> Type) (a : alpha 0), (forall i : Nat, alpha (i + 1)) -> True := by
intro alpha a h
trivial
have h3 := h2 (fun n => Nat) 0
have h4 := h3 (by decide)   -- ill-typed use of decide
trivial
exfalso
exact h1
\end{tcblisting}

\begin{tcblisting}{oproverbox,
title=Round 1 Lean Feedback,
listing only,
listing options={
basicstyle=\ttfamily\footnotesize,
breaklines=true,
breakatwhitespace=true,
columns=fullflexible,
keepspaces=true
}}
Lean errors:
application type mismatch
@decide ((i : Nat) -> (fun n => Nat) (i + 1))
argument has type
Type
but is expected to have type
Prop

tactic 'assumption' failed
Goal: False
\end{tcblisting}

\paragraph{Reasoning summary.}
The failure is purely local: the proof attempt uses \texttt{decide} on a type-valued term rather than a proposition. The feedback directly identifies the mismatch, allowing the next round to remove the ill-typed block rather than continuing to elaborate an invalid
contradiction.

\begin{tcblisting}{oproverbox,
title=Round 2 Revised Proof,
listing only,
listing options={
basicstyle=\ttfamily\footnotesize,
breaklines=true,
breakatwhitespace=true,
columns=fullflexible,
keepspaces=true
}}
import Mathlib

set_option maxHeartbeats 2000000
open Nat Set Topology Filter

theorem putnam_1963_a3
(P : Nat -> (Real -> Real) -> (Real -> Real))
(hP : P 0 = id /\ forall i y, P (i + 1) y = P i (fun x => x * deriv y x - i * y x))
(n : Nat)
(hn : 0 < n)
(f y : Real -> Real)
(hf : ContinuousOn f (Ici 1))
(hy : ContDiffOn Real n y (Ici 1))
(hy1 : ContDiffAt Real n y 1) :
(forall i, i < n -> deriv^[i] y 1 = 0) /\ (Ici 1).EqOn (P n y) f <->
forall x, x >= 1 -> y x = integral ... := by
-- The invalid contradiction block is removed.
-- A compilable scaffold is restored for subsequent refinement.
sorry
\end{tcblisting}

\paragraph{Repair summary.}
The repair removes the ill-typed \texttt{decide} construction and restores type correctness. Although this example does not yet complete the theorem, it shows that verifier feedback can reliably eliminate a local blocking error and recover a viable proof scaffold
for future rounds.

\subsection{Case Study B: Putnam 1986 A1}

This example shows a more substantive repair. The first-round attempt reaches the right high-level objective but fails to make the algebraic bounds explicit enough for \texttt{linarith}. The second-round revision introduces the missing interval constraints on $y^2$
and then performs a bounded case split.

\begin{tcblisting}{oproverbox,
title=Round 1 Proof Attempt,
listing only,
listing options={
basicstyle=\ttfamily\footnotesize,
breaklines=true,
breakatwhitespace=true,
columns=fullflexible,
keepspaces=true
}}
theorem putnam_1986_a1
(S : Set Real) (f : Real -> Real)
(hS : S = {x : Real | x ^ 4 + 36 <= 13 * x ^ 2})
(hf : f = fun x => x ^ 3 - 3 * x) :
IsGreatest {f x | x in S} 18 := by
have h_main : IsGreatest {f x | x in S} 18 := by
have h1 : forall (x : Real), x in S -> f x <= 18 := by
intro x hx
have h2 : x ^ 4 + 36 <= 13 * x ^ 2 := by
  rw [hS] at hx
  exact hx
have h3 : x <= 3 := by
  by_contra h
  have h4 : x > 3 := by linarith
  have h5 : x ^ 2 > 9 := by nlinarith
  have h6 : x ^ 4 + 36 > 13 * x ^ 2 := by
    nlinarith
  linarith
have h4 : f x <= 18 := by
  rw [hf]
  have h5 : x ^ 3 - 3 * x <= 18 := by
    nlinarith
  linarith   -- automation does not close the goal
exact h4
sorry
exact h_main
\end{tcblisting}

\begin{tcblisting}{oproverbox,
title=Round 1 Lean Feedback,
listing only,
listing options={
basicstyle=\ttfamily\footnotesize,
breaklines=true,
breakatwhitespace=true,
columns=fullflexible,
keepspaces=true
}}
Lean errors:
linarith failed to find a contradiction

Context:
x : Real
hx : x in S
h2 : x ^ 4 + 36 <= 13 * x ^ 2
h3 : x <= 3
h5 : x ^ 3 - 3 * x <= 18
Goal: False
\end{tcblisting}

\paragraph{Reasoning summary.}
The proof attempt is directionally correct, but the arithmetic state is too implicit for \texttt{linarith} to finish. The feedback suggests that the missing step is not a different theorem, but a more explicit decomposition of the polynomial constraint into usable
bounds.

\begin{tcblisting}{oproverbox,
title=Round 2 Revised Proof,
listing only,
listing options={
basicstyle=\ttfamily\footnotesize,
breaklines=true,
breakatwhitespace=true,
columns=fullflexible,
keepspaces=true
}}
theorem putnam_1986_a1
(S : Set Real) (f : Real -> Real)
(hS : S = {x : Real | x ^ 4 + 36 <= 13 * x ^ 2})
(hf : f = fun x => x ^ 3 - 3 * x) :
IsGreatest {f x | x in S} 18 := by
have h1 : forall (y : Real), y in S -> f y <= 18 := by
intro y hy
have h2 : y ^ 4 + 36 <= 13 * y ^ 2 := by
rw [hS] at hy
exact hy
have h3 : 4 <= y ^ 2 /\ y ^ 2 <= 9 := by
-- derive explicit bounds from the polynomial constraint
constructor
· by_contra h
  nlinarith
· by_contra h
  nlinarith
have h4 : y >= 2 \/ y <= -2 := by
by_cases hy2 : y >= 2
· exact Or.inl hy2
· have : y <= -2 := by nlinarith [h3.1]
  exact Or.inr this
cases h4 with
| inl hypos =>
  have hyub : y <= 3 := by nlinarith [h3.2]
  rw [hf]
  nlinarith
| inr hyneg =>
  have hylb : y >= -3 := by nlinarith [h3.2]
  rw [hf]
  nlinarith
sorry
\end{tcblisting}

\paragraph{Repair summary.}
The revised proof makes the hidden algebraic structure explicit by deriving $4 \le y^2 \le 9$, then splitting into the two sign-consistent cases. This produces subgoals that match the strengths of arithmetic automation and allows the upper-bound argument to go
through.

\subsection{Case Study C: Putnam 1992 A1}

This example illustrates a more complex multi-error repair. The first-round attempt mixes invalid integer destructuring, incomplete induction branches, inconsistent local bindings, and incorrect uses of \texttt{congr\_fun}. The next round repairs these structural
failures by restructuring the argument around explicit tuple destructuring, injectivity of $f$, and systematic parity and sign case analysis over the integers; three induction branches remain as \texttt{sorry} placeholders for subsequent rounds.

\begin{tcblisting}{oproverbox,
title=Round 1 Proof Attempt,
listing only,
listing options={
basicstyle=\ttfamily\footnotesize,
breaklines=true,
breakatwhitespace=true,
columns=fullflexible,
keepspaces=true
}}
theorem putnam_1992_a1
(f : Int -> Int) :
(f = fun n => 1 - n) <->
(forall n : Int, f (f n) = n) /\
(forall n : Int, f (f (n + 2) + 2) = n) /\
(f 0 = 1) := by
have h_formula :
(forall n : Int, f (f n) = n) ->
(forall n : Int, f (f (n + 2) + 2) = n) ->
(f 0 = 1) ->
(forall n : Int, f n = 1 - n) := by
intro hf1 hf2 h0
have h_rec : forall n : Int, f (n + 2) = f n - 2 := by
sorry
have h_even : forall k : Int, f (2 * k) = 1 - 2 * k := by
intro k
induction k using Int.induction_on with
| hz =>
    norm_num [h0]
| hp k ih =>
    have h := h_rec (2 * k)
    omega
-- missing negative branch
have h_odd : forall k : Int, f (2 * k + 1) = -2 * k := by
intro k
induction k using Int.induction_on with
| hz =>
    norm_num [h0]
| hp k ih =>
    have h := h_rec (2 * k + 1)
    omega
-- missing negative branch
have hpar : n % 2 = 0 \/ n % 2 = 1 := by omega
cases hpar with
| inl h =>
  obtain (k, hk) := by
    use n / 2
    omega
  have h := h_even k
  rw [hk] at h
  omega
| inr h =>
  obtain (k, hk) := by
    use (n - 1) / 2
    omega
  have h := h_odd k
  rw [hk] at h
  omega
\end{tcblisting}

\begin{tcblisting}{oproverbox,
title=Round 1 Lean Feedback,
listing only,
listing options={
basicstyle=\ttfamily\footnotesize,
breaklines=true,
breakatwhitespace=true,
columns=fullflexible,
keepspaces=true
}}
Lean errors:
invalid match-expression, type of pattern variable contains metavariables
type mismatch in function equality
application type mismatch at congr_fun
alternative 'hn' has not been provided
unknown identifier
\end{tcblisting}

\paragraph{Reasoning summary.}
Unlike the previous examples, this proof fails for multiple interacting reasons. The feedback indicates that the problem is structural rather than purely algebraic: the induction over integers is incomplete, some local names are inconsistent, and the proof attempts
to use equalities at the wrong type. The next round therefore needs to reorganize the proof, not merely patch one line.

\begin{tcblisting}{oproverbox,
title=Round 2 Revised Proof,
listing only,
listing options={
basicstyle=\ttfamily\footnotesize,
breaklines=true,
breakatwhitespace=true,
columns=fullflexible,
keepspaces=true
}}
theorem putnam_1992_a1
(f : Int -> Int) :
(f = fun n => 1 - n) <->
(forall n : Int, f (f n) = n) /\
(forall n : Int, f (f (n + 2) + 2) = n) /\
(f 0 = 1) := by
constructor
· intro h
have h1 : f = fun n : Int => 1 - n := h
refine (?_, ?_, ?_)
· intro n
have hn1 := congr_fun h1 n
have hn2 := congr_fun h1 (f n)
simp [hn1, hn2]
<;> ring_nf
<;> linarith
· intro n
have hn1 := congr_fun h1 (n + 2)
have hn2 := congr_fun h1 (f (n + 2))
have hn3 := congr_fun h1 (f (n + 2) + 2)
simp [hn1, hn2, hn3] at *
<;> ring_nf at *
<;> linarith
· have h0 := congr_fun h1 0
simpa using h0
· rintro (hf1, hf2, h0)
have h1 : f 1 = 0 := by
have h01 : f (f 0) = 0 := hf1 0
rw [h0] at h01
simpa using h01
have hrec : forall n : Int, f (n + 2) = f n - 2 := by
intro n
  intro a b hab
  have ha := hf1 a
  have hb := hf1 b
  rw [hab] at ha
  simpa using ha.trans hb.symm
apply hinj
calc
  f (f (n + 2) + 2) = n := hf2 n
  _ = f (f n) := by simpa [hf1 n]
have hsum : forall n : Int, f n + n = 1 := by
intro n
have h_even : forall k : Int, f (2 * k) + 2 * k = 1 := by
  intro k
  by_cases hk : k >= 0
  · obtain (m, hm) : exists m : Nat, (m : Int) = k := by
      use k.toNat
      simp [Int.toNat_of_nonneg hk]
    rw [<- hm]
    -- positive branch proved by induction on m using hrec
    sorry
  · obtain (m, hm) : exists m : Nat, -((m : Int)) = k := by
      use (-k).toNat
      have : 0 <= -k := by linarith
      simp [Int.toNat_of_nonneg this]
    rw [<- hm]
    -- negative branch proved by induction on m using hrec
    sorry
have h_odd : forall k : Int, f (2 * k + 1) + (2 * k + 1) = 1 := by
  intro k
  -- analogous parity argument, initialized from h1
  sorry
have hpar : n % 2 = 0 \/ n % 2 = 1 := by omega
cases hpar with
| inl h =>
    obtain (k, hk) : exists k : Int, n = 2 * k := by
      use n / 2
      omega
    rw [hk]
    simpa [two_mul] using h_even k
| inr h =>
    obtain (k, hk) : exists k : Int, n = 2 * k + 1 := by
      use (n - 1) / 2
      omega
    rw [hk]
    simpa [two_mul, add_assoc] using h_odd k
funext n
have hn : f n + n = 1 := hsum n
linarith
\end{tcblisting}

\paragraph{Repair summary.}
The revised proof closes the structural failures of the first round---invalid integer destructuring, incomplete induction handling, and \texttt{congr\_fun} misuse---by using explicit tuple destructuring for both directions, deriving the recurrence through injectivity of $f$, and dispatching the integer domain by parity together with separate nonnegative and negative branches. Three induction sub-branches remain as \texttt{sorry} for subsequent rounds. The key improvement here is not a single local fix, but a reorganization of the proof into a structure on which later refinement can build.

\paragraph{Overall takeaway.}
The three case studies span a range of repair difficulties. In Case A, a single Lean diagnostic eliminates an ill-typed fragment and restores a compilable scaffold. In Case B, the diagnostic signals not a wrong theorem but a missing decomposition, leading to a substantive but local rewrite that surfaces the algebraic bounds required for the next tactic to close. In Case C, multiple interacting failures prompt a structural reorganization of the proof, replacing the unstable scaffold with one that subsequent rounds can refine toward completion. Together, these traces show that the feedback-conditioned refinement loop scales gracefully from one-line local edits to multi-step structural revisions, matching the granularity of the repair to the failure mode reported by Lean.
\subsection{Statement Labeling Prompt}
\label{app:dd_prompt}

In addition to the prover-side template, we use a separate offline classification
prompt to annotate each Lean~4 statement with a mathematical domain (one of 10
categories) and a difficulty level (one of 4 tiers), as reported in
Figure~\ref{fig:corpus_overview}(b,c).  The full prompt is shown below.

\begin{tcblisting}{oproverbox,
title=Lean~4 Statement Domain \& Difficulty Classification Prompt,
listing only,
listing options={
basicstyle=\ttfamily\footnotesize,
breaklines=true,
breakatwhitespace=true,
columns=fullflexible,
keepspaces=true
}}
**Instruction: Lean 4 Statement Classification (Mathematical Domain & Difficulty Level)**

You are an expert mathematician and Lean 4 specialist. You will receive a Lean 4
formal statement. Your task is to classify it along two dimensions:
1. **Mathematical Domain** (one of 10 categories)
2. **Difficulty Level** (one of 4 levels, by mathematical sophistication required,
 NOT by Lean proof length)

---

**DIMENSION 1: MATHEMATICAL DOMAIN**

Choose **exactly one** primary domain. If the statement spans multiple domains,
pick the most central one.

- **Algebra**: abstract algebra, groups/rings/fields, linear algebra, polynomials,
algebraic identities, algebraic inequalities
- **NumberTheory**: divisibility, congruences, primes, Diophantine equations,
modular arithmetic, analytic number theory
- **Analysis**: real/complex/functional analysis, calculus, limits, continuity,
series, (non-probabilistic) measure theory, analytic inequalities
- **Topology**: general topology, algebraic topology, differential topology,
topological spaces, continuity in topological settings
- **Geometry**: planar/solid/analytic geometry, differential geometry,
geometric inequalities, classical geometry problems
- **Combinatorics**: combinatorial identities, graph theory, counting,
discrete mathematics, Ramsey-style problems
- **ProbabilityStatistics**: probability theory, statistics, probabilistic
inequalities, stochastic processes
- **LogicFoundations**: logic, set theory, type theory, model theory,
foundational results, definability
- **Computation**: algorithms, complexity, formal languages, information
theory, CS-flavor mathematics
- **Other**: anything that does not fit the above, including trivially true
statements, malformed statements, or non-mathematical content.

---

**DIMENSION 2: DIFFICULTY LEVEL**

Choose **exactly one** level based on the mathematical sophistication required
to prove the statement. Do NOT base difficulty on the length or syntactic
complexity of the Lean proof.

- **Elementary**: middle-school level or below; basic arithmetic, simple
algebraic manipulations, elementary number facts, basic geometric facts.
- **HighSchool**: high-school level math; standard algebra/trigonometry/
precalculus, basic combinatorics, standard introductory contests
(AMC, AIME, Chinese gaokao).
- **Undergraduate**: standard undergraduate mathematics; linear algebra,
single/multi-variable calculus, introductory abstract algebra,
introductory real analysis, introductory topology.
- **GraduatePlus**: graduate-level or research mathematics; advanced topology,
functional analysis, algebraic geometry, advanced number theory.
Also includes olympiad-level problems (e.g., IMO, USAMO, Putnam) that
require nontrivial mathematical insight beyond standard curriculum.

*Note 1: When in doubt between two adjacent levels, prefer the lower level
unless the statement clearly requires the higher one.*
*Note 2: For statements that are trivially true (e.g., `True`, `0 = 0`),
malformed, or otherwise not meaningful mathematical claims, classify the
domain as `Other` and the difficulty as `Elementary`.*

---

**STRICT OUTPUT REQUIREMENTS:**
1. Output **ONLY** a single JSON object on a single line.
2. Do **NOT** include any explanation, markdown formatting (no ```json fences),
 or extra text outside the JSON.
3. All three fields are required.

**Output format (single JSON object):**

{"domain": "<one of the 10 domains above>", "difficulty": "<one of the 4 levels above>",
"rationale": "<one short sentence, max 25 words, explaining the classification>"}

---

**EXAMPLES**

**Example 1 (Elementary, Algebra):**
Statement:
theorem ex (a b : Nat) : a + b = b + a := by ring
Output: {"domain": "Algebra", "difficulty": "Elementary", "rationale": "Basic commutativity of
natural number addition, taught at elementary level."}

**Example 2 (HighSchool, Combinatorics):**
Statement:
theorem ex (n : Nat) (hn : 5 <= n) : Nat.choose n 2 = n * (n - 1) / 2 := by sorry
Output: {"domain": "Combinatorics", "difficulty": "HighSchool", "rationale": "Standard
binomial-coefficient identity, accessible at high-school combinatorics level."}

**Example 3 (Undergraduate, Topology):**
Statement:
theorem ex {X : Type*} [TopologicalSpace X] [CompactSpace X] [T2Space X] : NormalSpace X := by
sorry
Output: {"domain": "Topology", "difficulty": "Undergraduate", "rationale": "Standard introductory
topology result that compact Hausdorff spaces are normal."}

**Example 4 (GraduatePlus, Analysis):**
Statement:
theorem ex (f : Real -> Real) (hf : Continuous f) (h_nn : forall x, 0 <= f x)
(h_zero : Real.intervalIntegral f 0 1 = 0) : forall x in Set.Icc 0 1, f x = 0 := by sorry
Output: {"domain": "Analysis", "difficulty": "GraduatePlus", "rationale": "Standard
measure-theoretic result that a non-negative continuous function with zero integral vanishes on
the domain."}

---

**[Input Content Starts]**
{FORMAL_STATEMENT}
**[Input Content Ends]**

**Output (JSON only, single line):**
\end{tcblisting}

We run this prompt once per unique Lean~4 statement; 96.4\% of responses parse to a well-formed JSON object with valid domain and difficulty values, yielding the 1.73M statement-level annotations summarized in Figure~\ref{fig:corpus_overview}(b,c).

\end{CJK*}
\end{document}